\documentclass{article} % For LaTeX2e
\usepackage{iclr2025_conference,times}

\usepackage{arxiv}

\usepackage{hyperref}
\usepackage{url}            % simple URL typesetting
\usepackage{booktabs}       % professional-quality tables
\usepackage{amsfonts}       % blackboard math symbols
\usepackage{nicefrac}       % compact symbols for 1/2, etc.
\usepackage{microtype}      % microtypography
\usepackage{pifont}
\usepackage{adjustbox}
\usepackage{wrapfig}
\usepackage{multirow,mathtools}
\usepackage{rotating}
\usepackage{lscape}
\usepackage{longtable}
\usepackage{amssymb}
\usepackage[most]{tcolorbox}
\newtcolorbox{prompt}[2][]{colback=gray!5!white,colframe=gray!75!black,title=#2,#1}

\usepackage[font={footnotesize}]{caption}

\usepackage{multirow}
\usepackage{graphicx}
\graphicspath{{imgs/}}

\usepackage{wrapfig}

\usepackage{multirow,mathtools }

\usepackage{pifont}
\usepackage{color, colortbl}

\usepackage{blindtext}
\usepackage{lipsum}

\usepackage{multirow}
\usepackage{listings}

\usepackage{bbm}

\usepackage [english]{babel}
\usepackage[autostyle, english = american]{csquotes}

\usepackage{pifont}
\usepackage{url}
\usepackage[most]{tcolorbox}

\usepackage{lipsum}
\usepackage{soul}
\usepackage{xcolor}
\usepackage{wrapfig}
\usepackage{multirow,mathtools } 

\usepackage{adjustbox}
\MakeOuterQuote{"}

\usepackage{microtype}
\usepackage{graphicx}
\usepackage{subfigure}
\usepackage{booktabs} % for professional tables

\usepackage{hyperref}

\usepackage{amsmath}

\usepackage[capitalize,noabbrev]{cleveref}

\usepackage{nicefrac}       % compact symbols for 1/2, etc.

\usepackage{tablefootnote}

\usepackage{makecell}

\newcommand{\Def}[0]{\mathrel{\mathop:}=}

\usepackage{pifont}

\usepackage{color, colortbl}
\definecolor{Gray}{gray}{0.93}
\definecolor{Orange}{rgb}{1,0.5,0}
\definecolor{DGray}{gray}{0.83}
\definecolor{LightCyan}{rgb}{0.88,1,1}

\definecolor{WarnREd}{rgb}{1,0.4,0.4}
\definecolor{WarnOrange}{rgb}{1,0.682,0.502}
\definecolor{WarnPink}{rgb}{0.9176, 0.7215, 0.7215}
\definecolor{GoodGreen}{rgb}{0.5019, 0.9215, 0.6039}

\usepackage[T1]{fontenc}

\definecolor{styleblue}{HTML}{504099}
\definecolor{mypurple}{HTML}{9391ff}

\definecolor{bluegray}{rgb}{0.4, 0.6, 0.8}
\definecolor{ceruleanblue}{rgb}{0.16, 0.32, 0.75}

\hypersetup{
colorlinks=true,
citecolor=ceruleanblue,
linkcolor=ceruleanblue,
urlcolor=black
}

\definecolor{darkgreen}{rgb}{0.0, 0.45, 0.0}
\definecolor{darkred}{rgb}{0.5, 0.0, 0.0}
\definecolor{darkblue}{rgb}{0.0, 0.0, 0.5}
\definecolor{darkyellow}{rgb}{0.65, 0.65, 0}

\usepackage{float}

\usepackage{algorithm}    % For the floating environment
\usepackage{algorithmic}

\newcommand{\boaa}{{\bfseries\texttt{\textless|begin\_of\_audio|\textgreater}}}
\newcommand{\eoaa}{{\bfseries\texttt{\textless|end\_of\_audio|\textgreater}}}
\newcommand{\tokens}[1]{\textcolor{tokblue}{\texttt{[#1\ speech\ tokens]}}}
\newcommand{\texttokens}[1]{\textcolor{tokblue}{\texttt{[#1\ text\ tokens]}}} 
\definecolor{codebg}{HTML}{F6F8FA}
\definecolor{codeborder}{HTML}{D0D7DE}
\definecolor{tokblue}{HTML}{0969DA}
\usepackage[T1]{fontenc}
\usepackage[utf8]{inputenc}
\usepackage{xcolor}
\usepackage{tcolorbox}
\usepackage{courier} % 等宽字体（通用、稳定）。如需更现代外观可改用 inconsolata。

\usepackage{amsmath,amsfonts,bm}

\def\eqref#1{(\ref{#1})}
\def\1{\bm{1}}

\DeclareMathAlphabet{\mathsfit}{\encodingdefault}{\sfdefault}{m}{sl}
\SetMathAlphabet{\mathsfit}{bold}{\encodingdefault}{\sfdefault}{bx}{n}

\DeclareMathOperator*{\argmin}{arg\,min}

\DeclareMathOperator*{\minimize}{\text{minimize}}

\newcommand{\btheta}{{\boldsymbol{\theta}}}

\title{Subspace Control: Turning Constrained Model Steering into Controllable Spectral Optimization}

\author{
Yancheng Huang$^{1,*}$ ~~ 
Changsheng Wang$^{1,*}$ ~~ 
Chongyu Fan$^{1}$ ~~ 
Yicheng Lang$^{1}$ ~~ 
Bingqi Shang$^{1}$  \\
\textbf{Yang Zhang}$^{2}$ ~~ 
\textbf{Mingyi Hong}$^{3}$ ~~ 
\textbf{Qing Qu}$^{4}$ ~~ 
\textbf{Alvaro Velasquez}$^{5}$ ~~ 
\textbf{Sijia Liu}$^{1,2}$ \vspace*{1mm}\\
$^{1}$OPTML, Michigan State University, ~
$^{2}$MIT-IBM Watson AI Lab, IBM Research,\\
$^{3}$University of Minnesota, ~
$^{4}$University of Michigan, ~
$^{5}$University of Colorado Boulder \\
$^{*}$Equal contribution
}

\date{}
\begin{document}

\pagestyle{fancy}
\fancyhf{}
\cfoot{\thepage}

\maketitle

% \clearpage

\vspace*{-10mm}

 \begin{figure}[htb]
    \centering
    \vspace{-5mm}
    \includegraphics[width=0.9\linewidth]{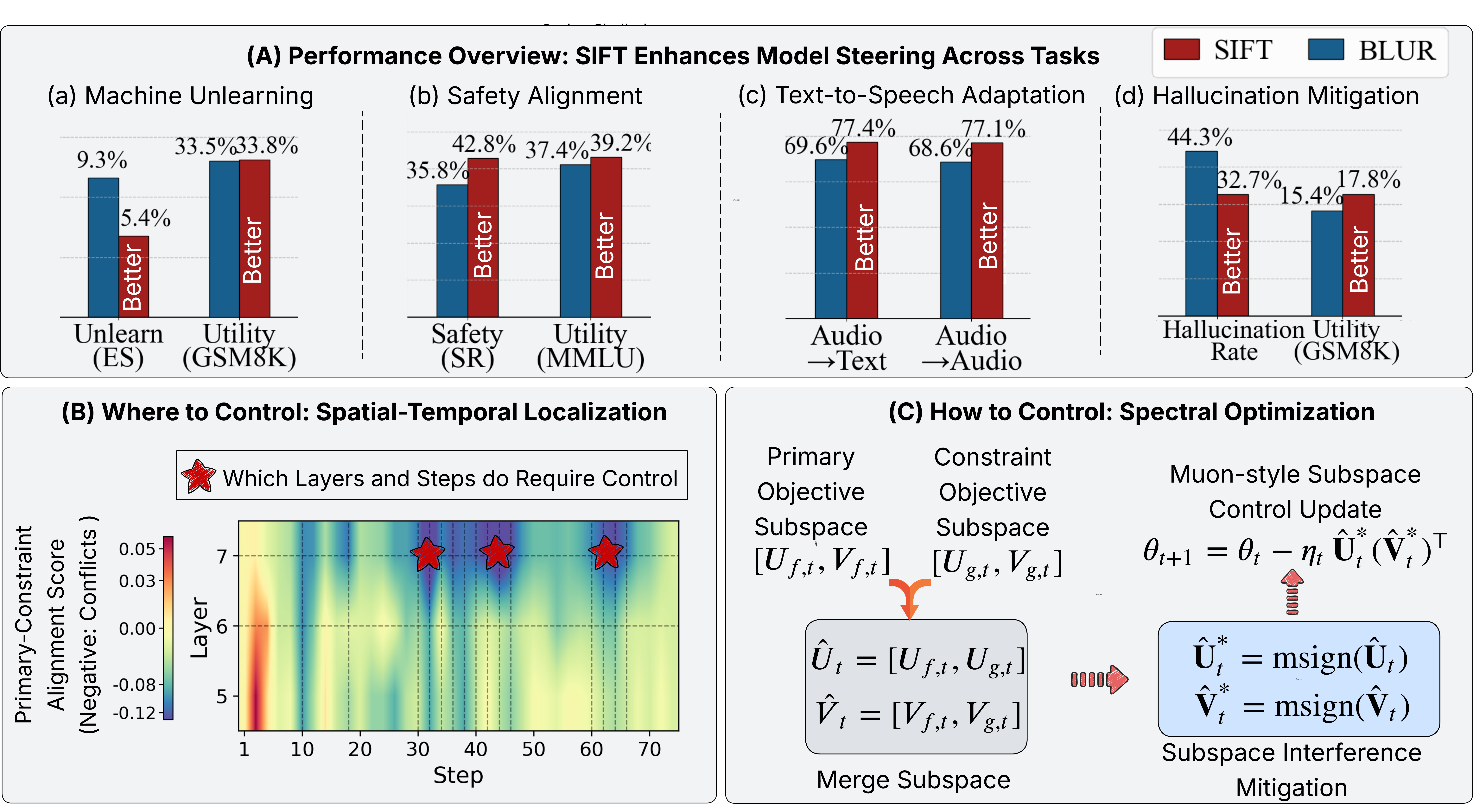}
    \vspace*{-2mm}
    \caption{{
    Schematic overview of proposed subspace control framework, SIFT. (\textbf{A}) Performance across four  model steering tasks (detailed in Tables\,\ref{tab:fg_spec} and \ref{tab:setup}), compared with the baseline BLUR \citep{reisizadeh2025blur}.
    %; in text-to-speech adaptation, “A→T” denotes \underline{A}udio-to-\underline{T}ext generation, and similarly for “A→A”. 
    (\textbf{B}) \textit{When and where to control}: SIFT enables selective intervention at targeted layers and training steps (\textit{i.e.}, spatial-temporal localization). (\textbf{C}) \textit{How to control}: Built on spectral optimizer Muon, SIFT leverages gradient orthogonalization (the matrix sign function) to mitigate subspace interference.
}}
\label{fig:overview}
\vspace*{-3mm}
\end{figure}

%\begin{abstract}

% \textbf{Abstract:}
% Foundation models, such as large language models (LLMs), are powerful but often require customization before deployment to satisfy practical constraints such as safety, privacy, and task-specific requirements, leading to ``constrained'' optimization problems for model steering and adaptation. 
% However, solving such problems remains largely underexplored and is particularly challenging due to interference between the primary objective and constraint objectives during optimization. 
% In this paper, we propose a subspace control framework for constrained model training. Specifically, (i) we first analyze, from a model merging perspective, how spectral cross-task interference arises and show that it can be resolved via a one-shot solution that orthogonalizes the merged subspace; (ii) we establish a connection between this solution and gradient orthogonalization in the spectral optimizer Muon; and (iii) building on these insights, we introduce \textit{SIFT} (spectral interference-free training), which leverages a localization scheme to selectively intervene during optimization, enabling controllable updates that mitigate objective–constraint conflicts. We evaluate SIFT across four representative applications: (a) machine unlearning, (b) safety alignment, (c) text-to-speech adaptation, and (d) hallucination mitigation. Compared to both control-based and control-free baselines, SIFT consistently achieves substantial and robust performance improvements across all tasks. 

% \textbf{Code:} \href{https://github.com/OPTML-Group/SIFT}{https://github.com/OPTML-Group/SIFT}.

 \begin{tcolorbox}[
  colback=white,
  % colframe={rgb:red,0;green,150;blue,120},
  colframe={rgb:red,24;green,69;blue,59},
  boxrule=0.6pt,
  arc=6pt,
  left=6pt,
  right=6pt,
  top=3pt,
  bottom=3pt
]
\textbf{Abstract:}
Foundation models, such as large language models (LLMs), are powerful but often require customization before deployment to satisfy practical constraints such as safety, privacy, and task-specific requirements, leading to ``constrained'' optimization problems for model steering and adaptation.
However, solving such problems remains largely underexplored and is particularly challenging due to interference between the primary objective and constraint objectives during optimization.
In this paper, we propose a subspace control framework for constrained model training. Specifically,
(i) we first analyze, from a model merging perspective, how spectral cross-task interference arises and show that it can be resolved via a one-shot solution that orthogonalizes the merged subspace;
(ii) we establish a connection between this solution and gradient orthogonalization in the spectral optimizer Muon; and
(iii) building on these insights, we introduce \textit{SIFT} (spectral interference-free training), which leverages a localization scheme to selectively intervene during optimization, enabling controllable updates that mitigate objective–constraint conflicts.
We evaluate SIFT across four representative applications: (a) machine unlearning, (b) safety alignment, (c) text-to-speech adaptation, and (d) hallucination mitigation.
Compared to both control-based and control-free baselines, SIFT consistently achieves substantial and robust performance improvements across all tasks.

% \vspace{0.5em}
\textbf{Code:} \href{https://github.com/OPTML-Group/SIFT}{https://github.com/OPTML-Group/SIFT}

\textbf{Correspondence:} \{huang341, wangc168, liusiji5\}@msu.edu
\end{tcolorbox}
\vspace{-5mm}

%\end{abstract}

%  \begin{figure}[htb]
%     \centering
%     \includegraphics[width=0.95\linewidth]{figure/HY/intro_v8.pdf}
%     \caption{{
%     Schematic overview of our subspace control framework, SIFT. (\textbf{A}) Performance across four  model steering tasks (see Tables\,\ref{tab:fg_spec} and \ref{tab:setup}), compared with the baseline BLUR \citep{reisizadeh2025blur}.
%     %; in text-to-speech adaptation, “A→T” denotes \underline{A}udio-to-\underline{T}ext generation, and similarly for “A→A”. 
%     (\textbf{B}) \textit{When and where to control}: SIFT enables selective intervention at targeted layers and training steps (\textit{i.e.}, spatial-temporal localization). (\textbf{C}) \textit{How to control}: Built on the spectral optimizer Muon, SIFT employs gradient orthogonalization (the matrix sign function $\mathrm{msign}$) to mitigate subspace interference.
% }}
% \label{fig:overview}
% \vspace{-5mm}
% \end{figure}

\clearpage
\newpage

\tableofcontents
% \newpage

% \SL{[Consider to make a highlighted introductory figure below abstract.]}

\section{Introduction}
\label{sec:intro}

Foundation models have achieved remarkable success across a wide range of applications. However, practical deployment \textit{rarely} occurs in an \textit{unconstrained} setting. Instead, pretrained models must be adapted to satisfy additional requirements. This naturally leads to \textit{constrained model steering} problems, where a primary objective (\textit{e.g.}, preserving general utility) must be optimized alongside additional constraint objectives, \textit{e.g.}, safety alignment \citep{ji2025pku, huang2503safety}, knowledge editing or removal \citep{meng2022locating,li2024wmdp}, or cross-modal adaptation beyond text \citep{cuervo2025closing,zeng2024scaling}.

% Despite its importance, existing approaches for solving such constrained model steering problems remain largely {superficial}, leaving the underlying challenges insufficiently understood. 
% This limitation is evidenced by the consistently observed \textit{poor trade-off} between optimizing the primary objective and satisfying the constraint objectives.
Despite its importance, existing approaches to constrained model steering remain under-explored, leaving the underlying challenges poorly understood. This is reflected in the consistently \textit{poor trade-off} between optimizing the primary objective and satisfying constraints.
For example, in the context of LLM unlearning for removing harmful knowledge generation capabilities \citep{shi2024muse,li2024wmdp,liu2025rethinking}, it has been recently observed that enforcing unlearning requirements can significantly degrade the model’s original instruction-following abilities \citep{fan2025llm}. 
% This degradation is not surprising. In the literature, 
Most existing model steering approaches approximate constraints via regularization or preference optimization \citep{ji2025pku,li2024wmdp,zhang2024negative}, effectively converting the problem into an ``unconstrained'' formulation for ease of optimization. However, such formulations often neglect conflicting optimization directions between the primary and constraint objectives, thereby inducing undesirable trade-offs in performance. 
% A central premise of this work is that the lack of effectiveness observed in constrained model steering is not use-case-specific, but instead stems from underlying \textit{algorithmic and structural limitations}.
% Such trade-off challenges observed in many model steering use cases reflect a common underlying algorithmic and structural limitation: as our empirical evidence shows that gradients induced by different objectives exhibit systematic and localized misalignment, suggesting the need for more targeted and controllable optimization interventions within current model steering algorithms. 
Such trade-off challenges, observed in many model steering use cases, reflect a common underlying \textit{algorithmic and structural limitation}.
As we will provide evidence, gradients induced by different objectives exhibit systematic and localized misalignment, highlighting the need for more targeted and controllable optimization interventions in modern model steering algorithms.
This observation raises a fundamental question:
\begin{tcolorbox}[before skip=2mm, after skip=0.0cm, boxsep=0.0cm, middle=0.0cm, top=0.1cm, bottom=0.1cm, boxrule=0.6pt]
\begin{center}
\textit{\textbf{(Q)}  
How can we develop principled and controllable optimization strategies to resolve structured and localized conflicts between primary and constraint objectives in model steering?
}
\end{center}
\end{tcolorbox}
\vspace*{2mm}

% In this paper, we answer \textbf{(Q)}  by introducing a \textit{subspace control} perspective that transforms constrained model steering into a problem of \textit{controllable spectral optimization}. 

To tackle  \textbf{(Q)}, we introduce a \textit{subspace control} perspective that transforms constrained model steering into a problem of \textit{controllable spectral optimization}.
Our approach is motivated by a {task interference perspective}, which establishes a novel connection between two seemingly distinct paradigms:
(i) \textit{one-shot model merging} \citep{gargiulo2025task,marczak2025no}, where task interference arises from non-orthogonal subspaces, and
(ii) \textit{iterative spectral optimization} via Muon (MomentUm Orthogonalized by Newton--Schulz) \citep{jordan2024muon,liu2025muon}, where gradient orthogonalization (via the matrix sign function) naturally mitigates such interference.
This connection reveals that conflicts between objectives can be systematically addressed through \textit{subspace orthogonalization}, providing a principled foundation for controllable optimization. 

Building on the above, we propose \textbf{SIFT} (Spectral Interference-Free Training), a method that enables \textit{localized subspace interventions} during model steering; see \textbf{Fig.\,\ref{fig:overview}} for an overview. Instead of globally modifying gradients or discarding conflicting components, SIFT constructs an interference-aware spectral subspace and applies orthogonalization in a \textit{targeted and adaptive} manner (Fig.\,\ref{fig:overview}B-C). This enables joint optimization of the primary and constraint objectives while preserving informative descent directions from both, overcoming the limitations of existing approaches. 
% We demonstrate the effectiveness of SIFT across \textit{four representative model steering tasks}: machine unlearning, safety alignment, text-to-speech adaptation, and hallucination mitigation. Across all settings, SIFT consistently outperforms both control-free and control-based baselines, achieving improved trade-offs between utility and constraint satisfaction.
Our \textbf{contributions} are summarized below:

$\bullet$ \textit{From unconstrained to constraint-aware to controllable optimization.} 
We formulate model steering as a constrained optimization problem and reveal the structural origin of objective conflicts through gradient misalignment across both temporal and spatial dimensions.
% We formulate model steering as a constrained optimization problem and identify gradient misalignment across temporal and spatial dimensions as the fundamental cause of objective conflicts.

$\bullet$ \textit{Subspace control conceptualization.} We establish a novel link between model merging and spectral optimization, showing that objective interference can be mitigated via subspace orthogonalization, realized through gradient orthogonalization in Muon.

%subspace orthogonalization,   opening up a principled design space for subspace control mechanisms enabled by gradient orthogonalization in Muon.

$\bullet$ \textit{SIFT for localized subspace control.} We propose SIFT, a spectral optimization method built on  Muon that enables  subspace interventions for targeted interference-aware optimization.

$\bullet$ \textit{Extensive evaluation across applications.} We demonstrate the effectiveness of SIFT across four representative model steering tasks, machine unlearning, safety alignment, text-to-speech adaptation, and hallucination mitigation, achieving improved performance over both control-free and control-based baselines, as highlighted in {Fig.\,\ref{fig:overview}}A.

%\vspace{-5pt}
\section{Related Work}
\label{sec:related-works}

\noindent \textbf{Model steering and adaptation.}
Model steering and adaptation often need modifying pretrained foundation models to satisfy new requirements or behaviors beyond their original training \citep{yang2024low,  sinii2025steering}. In practice, such processes are often inherently constrained \citep{ji2025pku, huang2503safety, li2024wmdp,zhang2024negative,cuervo2025closing,zeng2024scaling}.
However, due to the difficulty of handling objective–constraint conflicts \citep{lin2024mitigating,siddiqui2026position}, existing post-training approaches, such as supervised fine-tuning \citep{anisuzzaman2025fine,zhang2024balancing}, reinforcement learning \citep{jia2025beyond,shakya2023reinforcement}, and parameter-efficient adaptation \citep{han2024parameter,xu2026parameter}, often fail to achieve an effective trade-off, as they treat constrained steering and adaptation as unconstrained training.
% A straightforward method to resolve the objective–constraint conflict is BLUR \citep{reisizadeh2025blur}, an important baseline in our experiments, which projects the primary objective gradient onto the subspace orthogonal to the constraint gradient.
% However, BLUR’s performance is limited since it discards some useful components,which will be discussed thoroughly in this paper.
BLUR \citep{reisizadeh2025blur} addresses this via bi-level optimization, projecting the primary gradient orthogonally to the constraint gradient to mitigate conflicts. Yet, its performance remains limited, as shown later. 

\noindent \textbf{Model editing in spectral space.}
% Unlike approaches that directly manipulate the parameter space, a growing body of recent work has increasingly focused on spectral editing to reduce cross-task interference \citep{zhang2026gss,zhang2025lori,zhu2025path,biswas2025cure}.
% Additionally, a growing body of research on model merging has found that editing in the spectral space is more effective than operating directly in the parameter space \citep{gargiulo2025task,marczak2025no,yao2026merging}.
% These methods typically use SVD-based subspace analysis to disentangle task-specific subspaces, thereby minimizing interference across tasks.
Recent work increasingly focuses on spectral editing to reduce cross-task interference \citep{zhang2026gss,zhang2025lori,zhu2025path,biswas2025cure}, and studies on model merging show that editing in spectral space is more effective than direct parameter-space manipulation \citep{gargiulo2025task,marczak2025no,yao2026merging}.
One spectral editing method relevant to our work is POME \citep{liu2025pome}, which applies a Muon-style truncated SVD projection for model editing. However, it focuses on enhancing a fine-tuned model without a cross-task setting. Unlike existing model editing methods, our goal is not a one-shot fix but a more principled optimization approach. 
% in an iterative manner

\noindent \textbf{Spectral optimization and Muon.}
% The spectral optimizer Muon \citep{jordan2024muon} and its variants \citep{liu2025muon, ma2024swan,riabinin2025gluon, he2025low} have recently gained significant attention for improving fine-tuning stability and efficiency through gradient orthogonalization. This operator replaces the raw gradient with a geometrically well-conditioned one that emphasizes weaker directions. It can be viewed as a steepest-descent step under a spectral-norm constraint \citep{bernstein2024old}. In this paper, we establish a novel connection between one-shot model merging and Muon, showing that gradient orthogonalization can naturally reduce cross-task interference.
The Muon optimizer \citep{jordan2024muon} has recently become a notable spectral optimizer designed to improve training stability and efficiency through gradient orthogonalization. 
% which replaces raw gradients with geometrically well‑conditioned updates that emphasize weaker spectral directions. 
Empirically, Muon has achieved substantial improvements in \textit{pre}training efficiency and effectiveness across vision and language models 
%outperforming AdamW in both sample efficiency and wall‑clock time
\citep{jordan2024muon,liu2025muon,ma2026preconditioning}. 
% A growing body of research has also focused on understanding the mechanisms underlying Muon’s effectiveness \citep{jordan2024muon,chen2025muon,boreiko2025towards}. 
% and, more recently, an increasing number of variants have been proposed \citep{liu2025muon, ma2024swan,riabinin2025gluon, he2025low,kovalev2025understanding}.
% In contrast to this prior work, our approach focuses on the fine-tuning stage, with a particular emphasis on the constrained model steering problem.
Recent work has examined the mechanisms behind Muon’s effectiveness \citep{jordan2024muon,chen2025muon,boreiko2025towards}, with a primary emphasis on pretraining \citep{liu2025muon,ma2024swan,riabinin2025gluon,he2025low,kovalev2025understanding}. Although our method builds upon Muon, it differs from prior works by focusing on mitigating objective–constraint conflicts in the constrained model steering during the post-training stage.

\section{Formulation and Motivation: A Constraint-to-Control Perspective}
\label{sec:problem}

In this section, we first present a  \textit{constrained} optimization formulation, incorporating either hard (explicit) or soft (implicit) constraints, to steer a pre-trained model (\textit{e.g.}, LLMs) to satisfy requirements such as safety, privacy, and task-specific adaptation. We next provide a brief overview of how this formulation arises across representative use cases of our interest. Furthermore, we highlight the optimization challenges through a gradient alignment perspective, motivating the need for  more controllable optimization interventions to effectively solve such constrained problems.

\noindent \textbf{Problem formulation: Model training with ``constraints''.}
% Model steering typically entails optimizing structured objectives that preserve the original properties of the model (\textit{e.g.}, standard utility) while encouraging new capabilities  (\textit{e.g.}, safety, privacy, and task-specific adaptation). 
Model steering typically entails optimizing structured objectives that  preserve the original model properties
while encouraging new capabilities.
%\citep{lu2025alignment,liu2025rethinking}.
% Model steering typically involves a training process with structured optimization objectives to promote new capabilities (\textit{e.g.}, safety alignment \citep{lu2025alignment}) while preserving original model properties (\textit{e.g.}, standard utility \citep{liu2025rethinking}). 
%This yields a constrained training problem in which \textit{a primary objective} $f$ is optimized over a constrained space defined by \textit{another constraint objective} $g$. 
This leads to a \textit{constrained optimization problem} in which the primary objective $f$ is minimized under constraints defined by another objective $g$.
In practice, this is often expressed as a \textit{regularized optimization} problem that balances $f$ and $g$ during training. Thus, it can be formulated in either a \textit{hard}-constrained form or a \textit{soft}-regularized form:
\begin{align}
& \textbf{Hard-constrained:}  \quad \displaystyle \minimize_{\btheta \in {\Theta}}  ~f(\btheta)      \quad  \text{subject to} ~~  \Theta = \argmin_{\btheta} g(\btheta),
%\tag{hard formulation} 
\label{eq:hard_constraint}\\
 & \textbf{Soft-regularized:}  \quad ~~  \displaystyle \minimize_{\btheta} ~ \lambda f(\btheta) + g(\btheta) ~~~ \text{given regularization parameter $\lambda > 0$}.  
 %\tag{soft formulation}
 \label{eq:soft_constraint}
\end{align}
In \eqref{eq:hard_constraint}, the hard-constrained formulation can also be viewed as a simple bi-level optimization problem \citep{dempe2021simple,zhang2024introduction}, where the lower-level variables ($\btheta$) coincide with the upper-level variables, and the lower-level problem defines a solution set ($\Theta$) over which the upper-level objective is optimized. This formulation has been used to \textit{machine unlearning for LLMs} \citep{reisizadeh2025blur}, where the goal is to remove LLMs' unwanted capabilities while preserving useful ones. In this context, $f$ corresponds to the standard training loss for utility retention, while $g$ encodes the unlearning objective \citep{liu2025rethinking}.
% \SL{A natural extension of \eqref{eq:hard_constraint}–\eqref{eq:soft_constraint} is to consider multiple objectives beyond the primary and constraint objectives. Nevertheless, our focus in this work is not on multi-objective optimization.} 
%Additionally, the soft-regularized formulation \eqref{eq:soft_constraint} provides an alternative to \eqref{eq:hard_constraint} by reducing the bi-level problem to a single-level objective for ease of optimization, balancing the dual objectives via a regularization parameter $\lambda > 0$. 
%This formulation has also been used in machine unlearning, safety alignment, and model adaptation.

In both the hard- and soft-constrained settings \eqref{eq:hard_constraint}--\eqref{eq:soft_constraint}, we collectively refer to them as the \textit{constrained formulation}. 
A natural extension is to consider multiple objectives beyond the primary and constraint objectives; however, our focus and use cases in this work are \textit{not} on multi-objective optimization. 
\textbf{Table\,\ref{tab:fg_spec}} summarizes how the objectives $f$ and $g$ are specified across the applications considered in this work.

% {In both the hard- and soft-constrained settings \eqref{eq:hard_constraint}–\eqref{eq:soft_constraint}, we refer to them collectively as the       ``constrained formulation''.} 
% A natural extension is to consider multiple objectives beyond the primary and constraint objectives. Nevertheless, our focus and use cases in this work are \textit{not} on multi-objective optimization.
% In \textbf{Table~\ref{tab:fg_spec}}, we summarize how the objectives $f$ and $g$ are specified across the applications considered in this work.
%%%%%%%%%%% Prof's original Table %%%%%%%%%%

% \begin{table}[htb]
% \centering
% \caption{Specification of the constrained formulation across different applications.\SL{[You can add equation here or loss name with citation.]}}
% \label{tab:fg_spec}
% \resizebox{0.99\linewidth}{!}{
% \begin{tabular}{lll}
% \toprule
% \textbf{Application} & \textbf{Primary Objective $f$} & \textbf{Constraint Objective $g$} 
% %& \textbf{Remarks} 
% \\
% % \midrule
% % General Form & $f(\theta)$ & $g(\theta)$ & $\lambda$ balances trade-off \\
% \midrule
% Machine unlearning (\S\ref{sec:xxx}) & \SL{[You can add equation here or loss name with citation.]} &  %& 
% \\
% Safety alignment (\S\ref{sec:xxx}) &  & % &  
% \\
% Hallucination mitigation (\S\ref{sec:xxx}) &  &  %& 
% \\
% Text-to-speech adaptation (\S\ref{sec:xxx}) &  &  % &  
% \\
% \bottomrule
% \end{tabular}
% }
% \end{table}

%%%%%%%%%%% End of the Table %%%%%%%%%%
\begin{table}[htb]
\centering
\caption{{Specification of the constrained formulation across different applications.}}
\label{tab:fg_spec}
\resizebox{0.99\linewidth}{!}{
\begin{tabular}{lcc}
\toprule
\textbf{Application} & \textbf{Primary Objective $f$} & \textbf{Constraint Objective $g$} 
\\
\midrule
Machine unlearning (\S\ref{sec:unlearning_exp}) & \makecell[c]{MSE (mean squared error) loss of representation\\alignment on a retain dataset~\citep{li2024wmdp}} & \makecell[c]{
RMU (representation misdirection unlearning) \\loss on a forget dataset \citep{li2024wmdp}
%Drive representations of forget data toward \\ a random vector (e.g., RMU loss~\citep{li2024wmdp})
}  %& 
\\
\midrule
Safety alignment (\S\ref{sec:safety_exp}) & \makecell[c]{
CE (cross-entropy)-based SFT (supervised\\fine-tuning) loss on a utility dataset}
%{Supervised alignment on retain data \\ (e.g., SFT loss~\citep{rafailov2023direct})} 
& \makecell[c]{
DPO (direct preference optimization) loss \\on a safety dataset \citep{rafailov2023direct}
  } % &  
\\
\midrule
Text-to-speech adaptation (\S\ref{sec:speech_exp}) &\makecell[c]{CE loss on text generation \citep{zeng2024scaling}
}    &  \makecell[c]{CE loss on speech generation \citep{zeng2024scaling}
} % &  
\\
\midrule
Hallucination mitigation (\S\ref{sec:hallucination_exp}) &  \makecell[c]{
CE-based prediction loss on\\non-hallucinated tokens
} &  \makecell[c]{Negative CE loss on  hallucinated tokens
}%& 
\\
\bottomrule
\end{tabular}
}
\end{table}

\noindent \textbf{Motivation for controlled optimization: A gradient alignment challenge.}
Effectively solving the constrained
problems \eqref{eq:hard_constraint}–\eqref{eq:soft_constraint} is highly nontrivial, as the primary objective $f$ (\textit{e.g.}, utility loss) and the constraint objective $g$ (\textit{e.g.}, unlearning loss) often induce conflicting optimization directions due to their inherently different goals \citep{lin2024mitigating,siddiqui2026position}. 
%For example, the known ``alignment tax'' phenomenon~\citep{lin2024mitigating,siddiqui2026position} shows that alignment may incur a performance cost on tasks unrelated to alignment.
A direct way to characterize this conflict is by examining the alignment between the gradients 
of $f$ and $g$ during optimization, referred to as \emph{gradient alignment}. This can be quantified via the cosine similarity 
$\tau \Def \frac{(\nabla_{\btheta} f)^\top \nabla_{\btheta} g }{\| \nabla_{\btheta} f \|_2 \| \nabla_{\btheta} g \|_2} $, 
where $\nabla_{\btheta}$ denotes the gradient operator with respect to $\btheta$, and $\|\cdot\|_2$ is the $\ell_2$ norm.
If $\tau < 0$, it 
indicates gradient \textit{mis}alignment, where optimizing one objective comes at the expense of the other.

%\begin{wrapfigure}{r}{0.4\textwidth}
\begin{figure}[htb]
    \centering
    \includegraphics[width=0.56\linewidth]{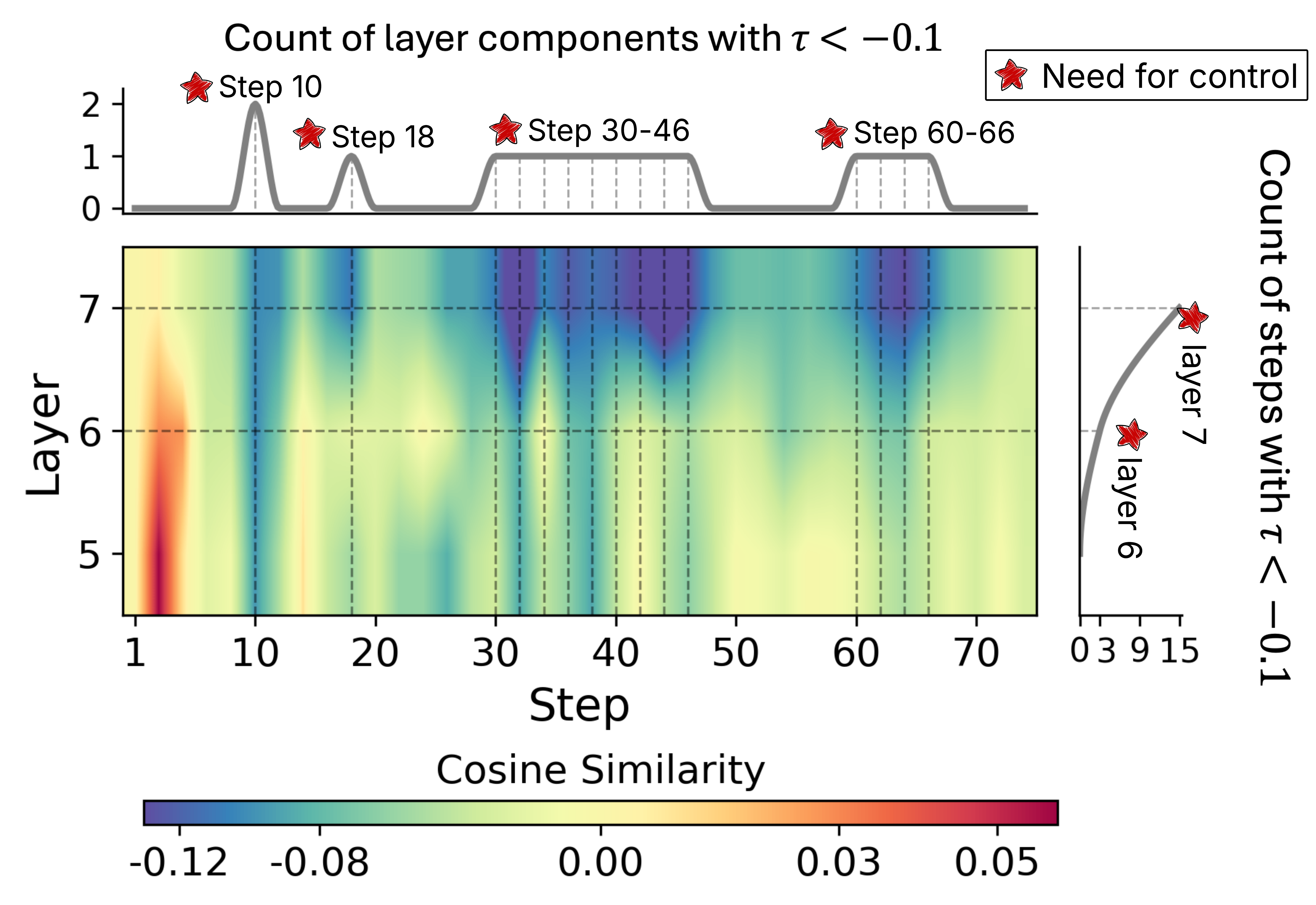}
\caption{{Visualization of cosine similarity $\tau$ across optimization steps (\textit{temporal} dimension) and model layers (\textit{spatial} dimension) in LLM unlearning.
%The heatmap shows the cosine similarity values $\tau$. 
The top and right marginal plots summarize the counts of $\tau < -0.1$ across steps and layers, respectively.
Red stars {\color{red}{$\star$}} mark  steps and layers need for control.}}
\label{fig:motivation_gradient_alignment}
%\vspace*{-5mm}
%\end{wrapfigure}
\end{figure}

\textbf{Fig.\,\ref{fig:motivation_gradient_alignment}} illustrates gradient misalignment across optimization steps (\textit{temporal}) and model layers (\textit{spatial}), using RMU 
(representation misdirection for unlearning) \citep{li2024wmdp} as a motivating example under the unlearning specification in Table~\ref{tab:fg_spec}. Experiments are conducted on the Zephyr-7B-Beta model with the WMDP dataset. Note that the original 
RMU implementation operates on layers 5, 6, and 7.
As we can see, negative values of $\tau$ 
persist at specific training steps and model layers, indicating a \textit{localized} pattern. In particular, conflicts are concentrated in higher layers (\textit{e.g.}, around layer 7) and occur across multiple 
training stages.
%We will later confirm that gradient misalignment arises in \textit{all} applications listed in Table~\ref{tab:fg_spec} and exhibits \textit{distinct localization patterns}. 
% The above motivates the development of a more controllable constrained optimization intervention that enables localized parameter updates during optimization to tolerate and mitigate misalignment between the primary and constraint objectives.
This motivates more controllable  optimization with localized updates to mitigate primary-constraint misalignment.
% The above motivates the development of a more controllable constrained optimization framework that can enable localized and principled interventions during optimization to tolerate and mitigate misalignment between the primary and constraint objectives.

 \begin{wrapfigure}{r}{0.25\textwidth}
    \centering
    \includegraphics[width=1\linewidth]{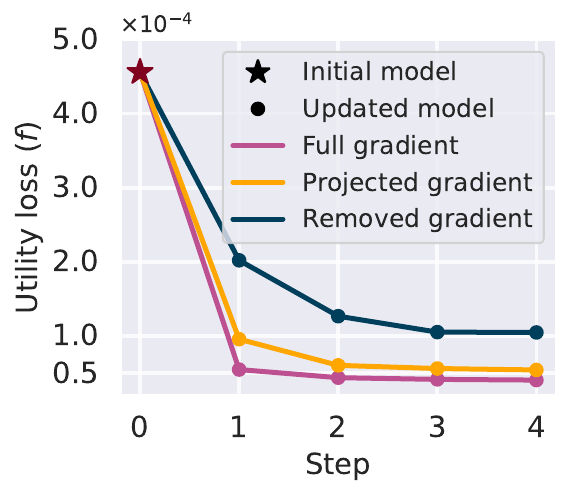}
    \vspace*{-6mm}
    \caption{{
Utility loss under different descent directions starting from a model at step 35 of the unlearning process. 
Multiple update steps are performed along the full gradient, projected gradient, and removed component, respectively. 
The unlearning setup is the same as in Fig.~\ref{fig:motivation_gradient_alignment}.
}}
\vspace*{-6mm}
\label{fig:limitation_gradient_projection}
\end{wrapfigure}
\noindent \textbf{A parameter-space optimization baseline: Gradient projection.}
% To mitigate gradient misalignment that complicates constrained optimization, a common approach constructs a descent direction based on the primary objective gradient $\nabla_{\btheta} f$ that avoids conflict with the constraint gradient $\nabla_{\btheta} g$. This can be achieved by using the component of $\nabla_{\btheta} f$ projected onto the subspace \emph{orthogonal} to $\nabla_{\btheta} g$, as in BLUR \citep{reisizadeh2025BLUR} for LLM unlearning.
To mitigate gradient misalignment, a common approach derives a descent direction from $\nabla_{\btheta} f$ that avoids conflict with $\nabla_{\btheta} g$ via orthogonal projection, as in BLUR \citep{reisizadeh2025blur}  for unlearning. That is,
\begin{align}
   \nabla_{\btheta} f^{\perp} \Def \left (\mathbf I - \mathbf G \right ) \nabla_{\btheta} f, ~~ \mathbf G \Def \frac{\nabla_{\btheta} g \,  (\nabla_{\btheta} g)^\top}{\|\nabla_{\btheta} g\|_2^2},
   \label{eq:grad_proj}
\end{align}
where $\mathbf I$ is the identity matrix, the term  $\mathbf G$ represents the subspace spanned by $\nabla_{\btheta} g$,  and thus its complement  $\mathbf I - \mathbf G$ 
represents the subspace orthogonal to $\nabla_{\btheta} g$.
Based on \eqref{eq:grad_proj}, we have $(\nabla_{\btheta} g)^\top \nabla_{\btheta} f^{\perp} = 0$. This implies that $\nabla_{\btheta} f^{\perp}$ is 
orthogonal to $\nabla_{\btheta} g$, ensuring {no negative gradient correlation}. 
Although gradient projection \eqref{eq:grad_proj} eliminates misalignment with the constraint objective $g$ by discarding the conflicting component of $\nabla_{\btheta} f$ (\textit{i.e.}, $\mathbf G \nabla_{\btheta} f$),  it does so at the cost of losing descent information for the primary objective $f$.
% \textbf{Fig.\,\ref{fig:limitation_gradient_projection}} in \textbf{Appendix\,\ref{app:gd_proj}} validates this limitation of gradient projection using the unlearning application in Fig.~\ref{fig:motivation_gradient_alignment}.
Under the unlearning setting specified in Table~\ref{tab:fg_spec} and Fig.~\ref{fig:motivation_gradient_alignment}, \textbf{Fig.\,\ref{fig:limitation_gradient_projection}} presents the model utility loss (encoded by $f$) across training steps when using the removed component $\mathbf{G}\nabla_{\boldsymbol{\theta}} f$, the full gradient $\nabla_{\boldsymbol{\theta}} f$, and the projected gradient $\nabla_{\boldsymbol{\theta}} f^{\perp}$ as descent directions, respectively.
% In \textbf{Fig.\,\ref{fig:limitation_gradient_projection}}, we compares the model utility loss (encoded by $f$) across training steps when using the removed component $\mathbf G \nabla_{\btheta} f$, the full gradient $\nabla_{\btheta} f$, and the projected gradient $\nabla_{\btheta} f^{\perp}$ as descent directions, respectively.
As we can see, the removed component $\mathbf G \nabla_{\btheta} f$ also contributes to reducing the utility loss. This indicates that discarding gradient components associated with one objective may waste useful descent information for the other objective.

\section{Our Method: Subspace Control}
\label{sec:spectral}

% In this section, we first introduce a  \textit{subspace control} perspective for constrained optimization in model steering and adaptation. To develop this view, we first present a simplified \textit{one-shot model merging} framework, illustrating how \textit{spectral interference} arises from non-orthogonal task subspaces. 
% We then connect model merging to the \textit{spectral optimization} framework Muon~\citep{jordan2024muon}, where \textit{gradient orthogonalization} (\textit{a.k.a.} matrix sign function) acts as a principled mechanism for eliminating such interference, serving as a control primitive in iterative optimization.

In this section, we introduce a \textit{subspace control} perspective for constrained optimization in model steering. We 
illustrate this via a \textit{one-shot model merging} framework, showing how \textit{spectral interference} arises from non-orthogonal task subspaces, and connect it to the \textit{spectral optimization} framework Muon~\citep{jordan2024muon}, where \textit{gradient orthogonalization} (\textit{a.k.a.} matrix sign function) provides a principled control primitive to eliminate such interference.
Building on that, we propose \textit{SIFT} (spectral interference-free training), a method that enables \textit{localized and controllable optimization interventions}. 

% within Muon by enforcing subspace orthogonality in the spectral domain.

\noindent \textbf{Spectral interference from task interaction: A model merging perspective.}
We adopt the task vector method~\citep{ilharco2022editing} to analyze the potential conflict between the primary and constraint objectives, yielding a simple one-shot model merging solution to \eqref{eq:hard_constraint}–\eqref{eq:soft_constraint}.
Specifically, consider two task vectors $\boldsymbol \Delta_f$ and $\boldsymbol \Delta_g$, defined as the parameter differences between the base model and the corresponding models fine-tuned on the primary and constraint objectives $f$ and $g$, respectively.
\textit{Task vector arithmetic} suggests that a merged model satisfying both objectives can be obtained by combining $\boldsymbol{\Delta}_f$ and $\boldsymbol{\Delta}_g$ (and applying the combined result $\boldsymbol \Delta$ to the base model): %\SL{[consider to add dimension to avoid confusion]}
\begin{align}
\boldsymbol \Delta \Def \boldsymbol \Delta_{f} + \boldsymbol \Delta_{g}
= %\underbrace{
\hat{\mathbf U}
%}_{\Def\, \hat U}
%\underbrace{
\begin{bmatrix}
\boldsymbol \Sigma_{f} & \mathbf{0} \\
\mathbf{0} & \boldsymbol \Sigma_{g}
\end{bmatrix}
\hat{\mathbf V}^\top
%}_{\Def\, \hat{\Sigma}}
%\underbrace{
,
\quad \hat{\mathbf U} \Def  [\mathbf U_{f},\, \mathbf U_{g}], 
\quad 
\hat{\mathbf V}  \Def  [\mathbf V_{f},\, \mathbf V_{g}]
%}_{\Def\, \hat V^\top}.
\label{eq:merge_tv}
\end{align}
where $\boldsymbol{\Delta}_f$ and $\boldsymbol{\Delta}_g$ admit \textit{compact} SVDs, with $\mathbf U_f$, $\mathbf U_g$ and $\mathbf V_f$, $\mathbf V_g$ denoting the left and right singular vector matrices, respectively, and $\boldsymbol{\Sigma}_f$, $\boldsymbol{\Sigma}_g$ the corresponding \textit{square} diagonal matrices of singular values with sizes determined by ranks.

However, direct merging in \eqref{eq:merge_tv} introduces \textit{``singular task interference''}, as noted in ~\citep{gargiulo2025task,marczak2025no}. That is, 
the combined bases $\hat{\mathbf U}$ and $\hat{\mathbf V}$ are generally non-orthogonal, \textit{i.e.}, $\hat{\mathbf U}^\top \hat{\mathbf U} \neq \mathbf I$ and $\hat{\mathbf V}^\top \hat{\mathbf V} \neq \mathbf I$, due to the non-orthogonality between $\mathbf U_f$ and $\mathbf U_g$ (and similarly $\mathbf V_f$ and $\mathbf V_g$). 
This  leads to \textit{spectral interference} across singular components.
To address it, a \textit{whitening transformation}  can be applied to  $\hat{\mathbf U}$ and $\hat{\mathbf V}$. This is equivalently formulated as an orthogonal Procrustes problem to seek orthogonal matrices $\mathbf U^*$ and $\mathbf V^*$ that are closest to $\hat{\mathbf U}$ and $\hat{\mathbf V}$ respectively \citep{{gargiulo2025task}}:
\begin{align}
\displaystyle \minimize_{\mathbf U} \;\; \bigl\lVert \mathbf U - \hat{\mathbf U} \bigr\rVert_F, \,\, \mathrm{s.t.} \;\; \mathbf U^\top \mathbf U = \mathbf I, \quad \text{yielding closed-form solution $\mathbf U^*   = \mathbf P \mathbf Q^\top$}.  
\label{eq:procrustes}
\end{align}
Here $\|\cdot\|_F$ denotes the Frobenius norm, and $\hat{\mathbf U}$ admits the compact SVD, with $\mathbf P$ and $\mathbf Q$ denoting the left and right singular vector matrices, respectively.
Replacing $\hat{\mathbf U}$ and $\hat{\mathbf V}$ with $\mathbf U^*$ and $\mathbf V^*$ in \eqref{eq:merge_tv} yields a \textit{non-interference} model merging. 
The key insight from the above model merging perspective is that spectral orthogonalization removes subspace interference, thereby reducing conflicts between the primary and constraint objectives and enabling their joint optimization in an interference-free subspace.

% It is also worth noting that, unlike gradient projection \eqref{eq:grad_proj}, which discards a gradient component to avoid task conflict, the spectral interference mitigation in \eqref{eq:procrustes} preserves both task subspaces $\mathbf U_f$ and $\mathbf U_g$ in $\mathbf U$ (similarly in $\mathbf V$) associated with the primary objective $f$ and the constraint objective $g$ respectively, while removing their interference.

\noindent \textbf{From model merging to Muon: Gradient orthogonalization as subspace control.}
Although model merging does \textit{not} provide an \textit{iterative} solver for \eqref{eq:hard_constraint}-\eqref{eq:soft_constraint}, a key observation from \eqref{eq:procrustes} is that the whitening transformation for interference mitigation aligns with the \textit{matrix sign function} ($\mathrm{msign}$), which serves as the \textit{gradient orthogonalization} step in the optimizer, \textit{Muon}. 
Muon can be interpreted as a steepest-descent method under a spectral-norm constraint~\citep{bernstein2024old}, yielding a principled spectral optimization framework that exploits the \emph{matrix-wise spectral structure} of descent directions rather than their {entry-wise} information.
%Next, we introduce the Muon algorithm and formalize its connection to task interference mitigation in \eqref{eq:procrustes}.
In Muon, given the iterate $\btheta_t$ at iteration $t$, the update to $\btheta_{t+1}$ is given by 
\begin{align}
\btheta_{t+1} = \btheta_t - \eta_t \, \mathrm{msign}(\mathbf M_t),
\label{eq:Muon}
\end{align}
where $\mathbf M_t$ denotes the current descent direction (\textit{e.g.}, gradient or momentum), and $\eta_t > 0$ is the step size. Compared to conventional optimizers such as SGD and Adam, the use of $\mathrm{msign}$ to perform gradient orthogonalization is a distinguishing feature of Muon, defined as follows (the iteration index $t$ is omitted for simplicity):
\begin{align}
\mathrm{msign}(\mathbf M)
= \boldsymbol{\Psi}\,\mathrm{sign}(\boldsymbol{\Sigma})\,\boldsymbol{\Phi}^\top
= \boldsymbol{\Psi}\, \boldsymbol{\Phi}^\top,
\label{eq:matrix_sign}
\end{align}
where $\mathbf M$ admits the compact SVD $\mathbf M = \boldsymbol{\Psi}\boldsymbol{\Sigma}\boldsymbol{\Phi}^\top$, with $\boldsymbol{\Psi}$ and $\boldsymbol{\Phi}$ being the left and right singular vector matrices, respectively, and $\boldsymbol{\Sigma}$ being the square diagonal matrix of singular values with size determined by its rank, and the function $\mathrm{sign}(\cdot)$ operates entry-wise, returning $1$ for diagonal singular values and $0$ for other entries in $\boldsymbol{\Sigma}$, \textit{i.e.}, {$\mathrm{sign}(\boldsymbol{\Sigma}) = \mathbf I$}.
Although SVD is used above, in practice $\mathrm{msign}(\mathbf M)$ is typically computed  via computationally-efficient Newton--Schulz iterations  \citep{jordan2024muon,liu2025muon}.

Comparing \eqref{eq:matrix_sign} with \eqref{eq:procrustes} yields several insights. \textit{First}, the descent direction $\mathbf M$ (\textit{e.g.}, momentum in our experiments) can be viewed as a generalized task vector capturing the difference between consecutive model updates. \textit{Second}, the role of gradient orthogonalization (\textit{i.e.}, $\mathrm{msign}$) in \eqref{eq:matrix_sign} parallels the subspace control in \eqref{eq:procrustes}, removing spectral interference when $\mathbf M$ contains components from the primary and constraint objectives.
%This leads to an important implication: \textit{Muon can serve as an effective optimizer for constrained problems} \eqref{eq:hard_constraint}–\eqref{eq:soft_constraint}.

\noindent \textbf{SIFT: Localized spectral control via Muon.}
Connecting model merging and Muon shows that Muon provides an algorithmic foundation for constrained optimization that mitigates spectral interference ``for free.''
Below, we formally introduce our proposed method, \textit{SIFT}.
The key idea behind SIFT is to construct an interference-free spectral subspace as in model merging,
% so that the primary objective $f$ and the constraint objective $g$  can be jointly optimized by Muon, 
with interference being naturally mitigated via $\mathrm{msign}$ \eqref{eq:matrix_sign}. The SIFT procedure is summarized in the following steps (a)-(d).

(a) We {first} obtain the momentum matrices $\mathbf{M}_{f,t}$ and $\mathbf{M}_{g,t}$ associated with the objectives $f$ and $g$ along the Muon optimization trajectory  at step $t$.

(b)  We {then} extract the \textit{top-$K$} spectral components of $\mathbf{M}_{f,t}$ and $\mathbf{M}_{g,t}$, denoted by $\mathbf{U}_{f,t}$ and $\mathbf{U}_{g,t}$ (and similarly $\mathbf{V}_{f,t}$ and $\mathbf{V}_{g,t}$). Similar to \eqref{eq:merge_tv}, these components are combined to form the expanded spectral subspaces:
\begin{align}
\hat{\mathbf{U}}_t = [ \mathbf{U}_{f,t}, \mathbf{U}_{g,t} ], ~~    \hat{\mathbf{V}}_t = [ \mathbf{V}_{f,t}, \mathbf{V}_{g,t} ].
\label{eq:expansion}
\end{align}
In our experiments, we find that $K$ can be set much smaller than the matrix dimension for some applications due to the low-rank structure of the momentum matrix; see \textbf{Fig.\,\ref{fig:topk_unlearn}} for validation.
Therefore, we treat $K$ as a hyperparameter for subspace curation.
%$\hat{\mathbf{U}}_t$ and $\hat{\mathbf{V}}_t$, respectively.

(c) We next apply the $\mathrm{msign}$ function in \eqref{eq:matrix_sign} to $\hat{\mathbf U}_t$ and $\hat{\mathbf{V}}_t$, computed via Newton–Schulz iterations, to obtain the interference-free subspaces $\hat{\mathbf{U}}_t^*$ and $\hat{\mathbf{V}}_t^*$, respectively:
\begin{align}
\hat{\mathbf{U}}_t^* = \mathrm{msign}(\hat{\mathbf{U}}_t), \quad
\hat{\mathbf{V}}_t^* = \mathrm{msign}(\hat{\mathbf{V}}_t).
\label{eq:msign_UV}
\end{align}
This step provides the key {controlled} optimization intervention for mitigating interference between the primary and constraint objectives $f$ and $g$ in the spectral subspace.

(d) We finally leverage $\hat{\mathbf{U}}_t^*$ and $\hat{\mathbf{V}}_t^*$ to construct a momentum-orthogonalized descent direction for updating the model parameters $\btheta$ in \eqref{eq:Muon}, yielding the \textbf{SIFT update}:
\begin{align}
    \btheta_{t+1} = \btheta_t - \eta_t \, \hat{\mathbf{U}}_t^* (\hat{\mathbf{V}}_t^*)^\top .
    %= \btheta_t - \eta_t \, \mathrm{msign}(\hat{\mathbf{U}}_t) \mathrm{msign}(\hat{\mathbf{V}}_t)^\top.
    \label{eq:SIFT_update}
\end{align}
The term $\hat{\mathbf{U}}_t^* (\hat{\mathbf{V}}_t^*)^\top$ can be interpreted as a gradient orthogonalization operator, formed from the interference-free, orthonormal subspaces of $\mathbf{M}_{f,t}$ and $\mathbf{M}_{g,t}$, as established in  \eqref{eq:expansion}--\eqref{eq:msign_UV}.

\begin{wrapfigure}{r}{0.3\textwidth}
\vspace*{-3mm}
    \centering
    \includegraphics[width=1\linewidth]{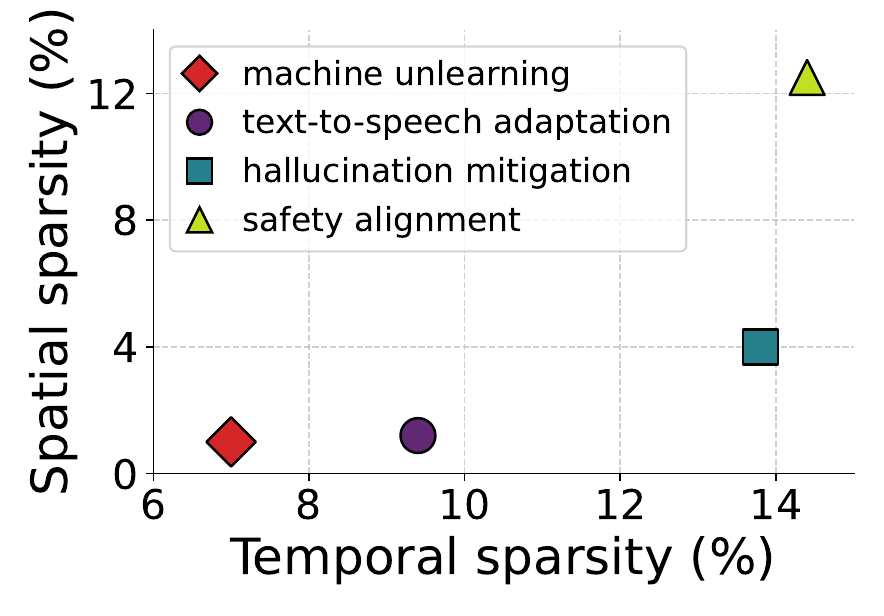}
    \vspace*{-5mm}
    \caption{
\textit{Sparse} localization across applications. 
Temporal sparsity is the fraction of all training steps, and spatial sparsity is the fraction of all model components where SIFT is activated.
}
    \label{fig:sparsity}
    \vspace*{-8mm}
\end{wrapfigure}
If we compare the  SIFT update  \eqref{eq:SIFT_update} with the standard Muon \eqref{eq:Muon}, SIFT (i) superposes the subspaces of $\mathbf{M}_{f,t}$ and $\mathbf{M}_{g,t}$ 
as in \eqref{eq:expansion}, and (ii) applies $\mathrm{msign}$ to the resulting expanded subspaces rather than to the overall momentum matrix, in order to mitigate spectral interference.
It is worth noting that, unlike gradient projection \eqref{eq:grad_proj}, which discards conflicting 
components, SIFT retains both task subspaces $\mathbf U_{f,t}$ and $\mathbf U_{g,t}$ in $\hat{\mathbf U}$ (and similarly in $\hat{\mathbf V}$),  while eliminating their interference via $\mathrm{msign}$ in \eqref{eq:msign_UV}.

In terms of computation, the SIFT update in \eqref{eq:SIFT_update} introduces additional overhead primarily due to SVDs in \eqref{eq:expansion}. However, 
this overhead is well justified, as SIFT is coupled with a \textit{localization} scheme, as measured by the gradient/momentum alignment score $\tau$.
This localization determines \textit{when} and \textit{where} SIFT should intervene within the standard Muon-based iterative optimization process, thereby enabling \textit{selective} application to specific optimization steps and model components, as indicated by {\color{red}{$\star$}} in Fig.\,\ref{fig:motivation_gradient_alignment}.
We emphasize that localization informs a \textit{sparse} subspace control pattern when applying SIFT across diverse applications. Fig.\,\ref{fig:sparsity} illustrates this sparsity over both temporal and spatial dimensions for different applications.
% This localization determines \textit{when} and \textit{where} SIFT should intervene
% as indicated by {\color{red}{$\star$}} in Fig.\,\ref{fig:motivation_gradient_alignment}.
% We emphasize that localization induces a \textit{sparse} subspace control pattern, intervening on fewer than 15\% of weights and steps across all applications, as shown in \textbf{Fig.\,\ref{fig:sparsity}}.
We refer readers to \textbf{Algorithm\,\ref{algo:sift}} in \textbf{Appendix\,\ref{app:alg}} for a detailed description of SIFT.

\section{Experiments}
\label{sec:experiment}

In this section, we evaluate the effectiveness of our proposed method, SIFT, against stateful baselines across four LLM steering applications listed in Table~\ref{tab:fg_spec}: machine unlearning, safety alignment,  text-to-speech adaptation, and hallucination mitigation.

\subsection{Experiment setups}

\noindent \textbf{Data-model-evaluation settings.}
As shown in \textbf{Table~\ref{tab:setup}}, we provide an overview of the experimental setups across the applications in Table~\ref{tab:fg_spec}, including the corresponding base models, datasets, and evaluation metrics. 
In our experiments, the base models and evaluation metrics are associated with the training datasets, following those specified in the corresponding benchmark releases. Unless otherwise noted, our applications focus on LLMs. For text-to-speech adaptation, we use GLM-4-Voice, an open-source speech language model that supports cross-modal input--output settings; for example, ``Audio $\rightarrow$ Text'' denotes audio input and text output.
In the safety alignment case, we use LLaMA-2-7B (rather than its chat variant) as the base model, as its weaker initial safety alignment makes it well suited for evaluating the effectiveness of subsequent alignment methods.
% Additional setup details for each application, along with some basic background, are provided in \textbf{Appendix\,\ref{app:setup}}.

% For the hallucination mitigation task, we adopt LLaMA-2-7B-Chat as the base model, as this choice is guided by the RAGTruth dataset, which provides a substantial number of hallucination-prone responses generated by this model. In the machine unlearning setting, we adopt Zephyr-7B as the base model, following the standard setup of the WMDP benchmark to enable direct comparison with prior work. In the safety alignment setting, we use LLaMA-2-7B as the base model, as it exhibits relatively weaker safety alignment, making it suitable for evaluating the effectiveness of further safety alignment methods.

\begin{table*}[htb]
\centering
\caption{Overview of experimental setups for different applications.}
\label{tab:setup}
\small
\resizebox{0.999\linewidth}{!}{
% \begin{tabular}{l l >{\raggedright\arraybackslash}m{7cm} >{\raggedright\arraybackslash}m{14cm} c}
\begin{tabular}{l l l l r}
\toprule
\textbf{Applications} & \textbf{Training Datasets} & \textbf{Base LLM} & \textbf{Evaluation}  \\ 
\midrule
% ====================================================
Machine unlearning
  & 
  \begin{tabular}[l]{@{}l@{}} 
  Forget: WMDP \citep{li2024wmdp}\\
   Retain:  Wikitext  \citep{merity2016pointer}  
  \end{tabular}       
    & zephyr-7b-beta
    & 
  \begin{tabular}[l]{@{}l@{}} 
  Unlearning ($\downarrow$): ES-Bio/Cyber \citep{yuan2024closer}, 
  MCQ-Bio/Cyber \citep{li2024wmdp}  \\
   Utility ($\uparrow$):   
   MMLU \citep{hendryckstest2021}, TruthfulQA \citep{lin2021truthfulqa}, \\
      IFEval \citep{zhou2023instructionfollowingevaluationlargelanguage}, 
      GSM8K \citep{cobbe2021gsm8k}  
  \end{tabular}     
    \\
     \midrule
% % ====================================================
Safety alignment
  & 
  \begin{tabular}[l]{@{}l@{}} 
 Safety: PKU-SafeRLHF  \citep{ji2024pku}\\
  Utility: Alpaca \citep{alpaca} 
  \end{tabular}       
    & Llama-2-7b
    & 
  \begin{tabular}[l]{@{}l@{}} 
 Safety ($\uparrow$) \citep{wang2025star1saferalignmentreasoning}:  
  Strong Reject, JBB-Behaviors, Wild Jailbreak \\
  Utility ($\uparrow$): 
  MMLU, GSM8K, IFEval,  MNLI \citep{wang2019glue}, MRPC \citep{wang2019glue}  
  \end{tabular}     
    \\
     \midrule
Hallucination mitigation
  & 
  \begin{tabular}[l]{@{}l@{}} 
RAGTruth \citep{niu2024ragtruth}
  \end{tabular}       
    & Llama-2-7B-Chat
    & 
  \begin{tabular}[l]{@{}l@{}} 
 Hallucination rate ($\downarrow$) \citep{niu2024ragtruth}  \\
  Utility ($\uparrow$): MMLU, GSM8K, TruthfulQA, QNLI
  \end{tabular}     
    \\
     \midrule
Text-to-speech adaptation
  & 
  \begin{tabular}[l]{@{}l@{}} 
ESNLI \citep{camburu2018snli}\\ COSE \citep{rajani2019explain}\\ OpenBookQA \citep{OpenBookQA2018}
  \end{tabular}       
    & 
    \begin{tabular}[l]{@{}l@{}} 
  GLM-4-Voice\\
  \citep{zeng2024glm}
  \end{tabular}     
    & 
  \begin{tabular}[l]{@{}l@{}} 
Test accuracy ({$\uparrow$}) on ESNLI, COSE, and OpenBookQA under four cross-modal input\\-output settings:   
“Audio $\rightarrow$ Text”, “Audio $\rightarrow$ Audio”, “Text $\rightarrow$ Audio”, “Text $\rightarrow$ Text”
  \end{tabular}     
    \\
\bottomrule
\end{tabular}
}
\end{table*}

\noindent \textbf{Baseline methods.} 
In all applications, we compare SIFT with \textbf{four} baseline methods \ding{172}--\ding{174} for steering a base model to meet the desired requirements.
Conventionally, model steering is typically solved as a regularized optimization problem \eqref{eq:soft_constraint} using a standard (control-free) optimizer such as \ding{172} \textbf{AdamW} \citep{loshchilov2017decoupled}. Since SIFT builds upon Muon, we also include the standard \ding{173} \textbf{Muon} \citep{jordan2024muon} as a baseline for solving the regularized problem \eqref{eq:soft_constraint}.
In addition, we include two baselines with explicit optimization interventions to handle conflicts between the primary and constraint objectives: \ding{174} \textbf{BLUR} \citep{reisizadeh2025blur} and \ding{175} \textbf{POME} \citep{liu2025pome}. 
% The former employs gradient projection as in \eqref{eq:grad_proj} to solve the hard-constrained problem \eqref{eq:hard_constraint}, while the latter adopts a model editing perspective, refining weight updates between pretrained and fine-tuned models via a single truncated SVD projection, similar to \eqref{eq:procrustes}. However, its one-shot design limits iterative adaptation and continuous improvement.

% In all application settings, we compare SIFT with two classical optimization methods without subspace control, AdamW and Muon, as well as two approaches with subspace control, POME and BLUR.
% POME refines the weight updates between pretrained and fine-tuned models via a single truncated SVD projection; however, its one-shot design limits iterative adaptation for continuous improvement. BLUR, on the other hand, employs the orthogonal projection described in \eqref{eq:grad_proj} to eliminate subspace interference. As shown in \textbf{Fig.\,\ref{fig:limitation_gradient_projection}} , this strategy may discard useful descent information.

\noindent \textbf{Implementation details.} 
To implement SIFT, there exist two key hyperparameters (Algorithm~\ref{algo:sift}): 
(a) the subspace dimension $K$ used to construct the spectral subspaces in \eqref{eq:expansion}, and 
(b) the misalignment threshold $\epsilon < 0$, which specifies when SIFT intervenes to mitigate interference between the primary and constraint objectives, as shown in Algorithm~\ref{algo:sift}.
% As shown in Algorithm~\ref{algo:sift}. SIFT performs subspace control when alignment score $\tau <\epsilon$.
For machine unlearning and safety alignment, we set $K=128$ and $K=192$, respectively, both significantly smaller than the matrix dimension (typically 4096 for 7B LLMs). 
This choice yields improved performance, consistent with the low-rank structure of the momentum matrices. In contrast, for text-to-speech adaptation and hallucination mitigation, we do not observe clear benefits from reducing $K$, and thus retain all components by default.
% As shown in Algorithm~\ref{algo:sift}, the threshold $\epsilon$  and is chosen case by case due to varying cosine similarity ranges across applications. 
The threshold $\epsilon$ is chosen case by case due to its varying cosine similarity ranges across applications.
Specifically, we set $\epsilon=-0.1$ for machine unlearning and safety alignment; for text-to-speech adaptation, $\epsilon=-0.6$ (ESNLI), $-0.4$ (COSE), and $-0.5$ (OpenBookQA); and for hallucination mitigation, $\epsilon=-0.8$.
All experiments are conducted on 8$\times$ NVIDIA A6000 GPUs with 48GB memory.

\subsection{Experiment results on LLM unlearning}
%\vspace*{-3mm}
\label{sec:unlearning_exp}
% In this application, we perform LLM unlearning using RMU on zephyr-7b-beta to remove its sensitive content generation capabilities on the forget set WMDP while preserving its general utility on the retain set Wikitext.
In this application, we perform LLM unlearning using representation misdirection for unlearning (RMU) \citep{li2024wmdp} on the base model zephyr-7b-beta to remove its sensitive content generation capabilities on the forget set (\textit{e.g.}, WMDP) while preserving general utility on the retain set (\textit{e.g.}, Wikitext). The specifications of the primary and constraint objectives, as well as the data--model--evaluation setups, are summarized in Table~\ref{tab:fg_spec} and Table~\ref{tab:setup}, respectively.

\begin{table}[htb]
\vspace*{-2mm}
\caption{Performance of LLM unlearning for 
% steering a base model to 
removing hazardous content generation on WMDP. ES-Bio/Cyber denotes the entailment score (ES)~\citep{yuan2024closer, fan2025llm} measuring unlearning effectiveness via factual consistency with the pre-unlearned model on WMDP Bio and Cyber sets. MCQ-Bio/Cyber denotes multi-choice question (MCQ) accuracy on the same sets; 
% lower values indicate better unlearning.
Model utility is evaluated on standard benchmarks.
% ({MMLU}, {TruthfulQA}, {IFEval}, {GSM8K}),
% where higher is better.
Superscripts indicate standard deviations over 10 random trials.}
\label{tab:sift_unlearn_main}
\centering
\small
\resizebox{0.999\linewidth}{!}{
\begin{tabular}{lcccc|cccc|c}
\toprule
\multirow{2}{*}{ \begin{tabular}[l]{@{}l@{}} 
 \textbf{Methods}
 \end{tabular}} 
& \multicolumn{4}{c}{\textbf{Unlearning Performance} (\%)} 
& \multicolumn{4}{c}{\textbf{Utility Performance} (\%)}
& \multirow{2}{*}{\makecell{\textbf{Runtime} \\ (min)}} \\
\cmidrule(lr){2-5} \cmidrule(lr){6-9}
& ES-Bio $\downarrow$ & ES-Cyber $\downarrow$ & MCQ-Bio $\downarrow$ & MCQ-Cyber $\downarrow$
& MMLU $\uparrow$ & TruthfulQA $\uparrow$ & IFEval $\uparrow$ & GSM8K $\uparrow$ & \\
\midrule
Base Model & 62.3 \scriptsize{$\pm 0.3$} & 46.7 \scriptsize{$\pm 0.2$} & 64.1 \scriptsize{$\pm 0.2$} & 44.8 \scriptsize{$\pm 0.4$} & 58.2 \scriptsize{$\pm 0.3$} & 39.5 \scriptsize{$\pm 0.2$} & 10.4 \scriptsize{$\pm 0.3$} & 35.6 \scriptsize{$\pm 0.4$} & N/A \\
\midrule
AdamW & 14.6 \scriptsize{$\pm 0.3$} & 26.3 \scriptsize{$\pm 0.2$} & 31.7 \scriptsize{$\pm 0.4$} & 29.4 \scriptsize{$\pm 0.3$} & 57.1 \scriptsize{$\pm 0.3$} & 40.8 \scriptsize{$\pm 0.4$} & 9.2 \scriptsize{$\pm 0.2$} & 33.9 \scriptsize{$\pm 0.3$} & 8.8 \\
Muon & 14.2 \scriptsize{$\pm 0.2$} & 30.5 \scriptsize{$\pm 0.3$} & 31.3 \scriptsize{$\pm 0.2$} & 28.6 \scriptsize{$\pm 0.2$} & 56.4 \scriptsize{$\pm 0.1$} & 37.9 \scriptsize{$\pm 0.2$} & 8.7 \scriptsize{$\pm 0.4$} & 34.1 \scriptsize{$\pm 0.3$} & 11.8 \\
POME & 15.8 \scriptsize{$\pm 0.4$} & 29.6 \scriptsize{$\pm 0.3$} & 29.2 \scriptsize{$\pm 0.3$} & 27.5 \scriptsize{$\pm 0.2$} & 57.3 \scriptsize{$\pm 0.3$} & 38.4 \scriptsize{$\pm 0.4$} & 9.6 \scriptsize{$\pm 0.2$} & 34.7 \scriptsize{$\pm 0.3$} & 9.9 \\
BLUR & 9.3 \scriptsize{$\pm 0.2$} & 24.8 \scriptsize{$\pm 0.2$} & 28.6 \scriptsize{$\pm 0.3$} & 27.3 \scriptsize{$\pm 0.3$} & 57.1 \scriptsize{$\pm 0.3$} & 39.2 \scriptsize{$\pm 0.1$} & 9.1 \scriptsize{$\pm 0.2$} & 33.5 \scriptsize{$\pm 0.3$} & 9.4 \\
\rowcolor{LightCyan!50}
SIFT & 5.4 \scriptsize{$\pm 0.3$} & 19.7 \scriptsize{$\pm 0.2$} & 26.8 \scriptsize{$\pm 0.2$} & 26.4 \scriptsize{$\pm 0.1$} & 56.8 \scriptsize{$\pm 0.3$} & 38.6 \scriptsize{$\pm 0.2$} & 10.3 \scriptsize{$\pm 0.3$} & 33.8 \scriptsize{$\pm 0.4$} & 20.2 \\
\bottomrule
\end{tabular}
}
\end{table}

% In this experiment, we perform LLM unlearning using representation misdirection for unlearning (RMU) \citep{li2024wmdp} on a base model to remove its sensitive content generation capabilities on the forget set (\textit{e.g.}, WMDP) while preserving general utility on the retain set (\textit{e.g.}, Wikitext). The specifications of the primary and constraint objectives, as well as the data--model--evaluation setups, are summarized in Table~\ref{tab:fg_spec} and Table~\ref{tab:setup}, respectively.

% \noindent \textbf{Performance and insights.}
% In \textbf{Table~\ref{tab:sift_unlearn_main}}, we present unlearning effectiveness (ES and MCQ on WMDP Bio/Cyber) and general utility ({MMLU}, {GSM8K}, {IFEval}, {TruthfulQA}), comparing SIFT with the original Zephyr-7B model and baseline methods (AdamW, Muon, POME, and BLUR).
In \textbf{Table~\ref{tab:sift_unlearn_main}}, we present unlearning effectiveness and general utility, comparing SIFT with the original Zephyr-7B model
and baseline methods.
As we can see, \textit{SIFT achieves the strongest unlearning performance on both multiple-choice (MCQ) and open-ended (ES) evaluations while maintaining competitive utility}. 
In particular, it attains $5.4\%$ ES-Bio and $19.7\%$ ES-Cyber, significantly outperforming BLUR ($9.3\%$ and $24.8\%$), 
% a baseline that explicitly uses gradient projection to mitigate the unlearning--utility conflict, 
as well as all other baselines. 
The larger gains on open-ended ES metrics indicate that SIFT more effectively removes underlying unwanted knowledge, rather than merely altering answer selection. Importantly, these improvements do not compromise utility: SIFT remains comparable to BLUR on benchmarks such as GSM8K (33.8\% vs.\ 33.5\%) and IFEval (10.3\% vs.\ 9.1\%). 

Compared to POME, which relies on one-shot task-vector-based intervention, SIFT performs multi-step, localized interference mitigation via spectral subspace control, resulting in much stronger unlearning (\textit{e.g.}, $5.4\%$ vs.\ $15.8\%$ on ES-Bio).
Notably, SIFT improves performance at the cost of increased runtime, \textit{e.g.}, roughly doubling that of the standard Muon optimizer. Improving its efficiency remains a direction for future work.

\begin{figure}[htb]
    \centering
    % \vspace*{-7mm}
    \begin{tabular}{cc}
    \includegraphics[width=0.32\textwidth]{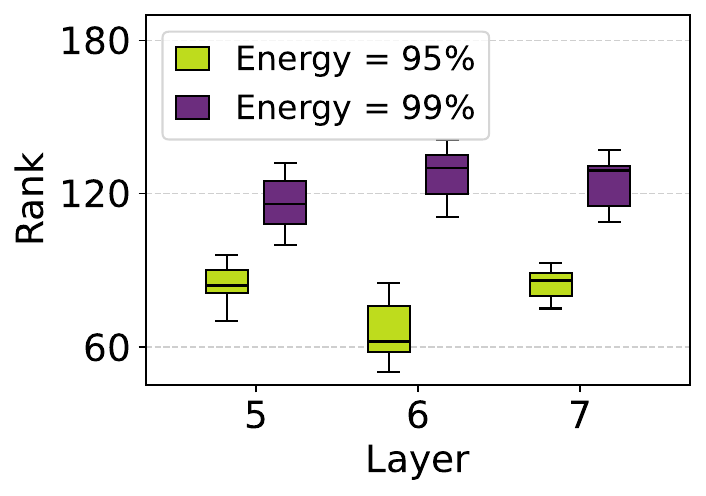}    &   \hspace*{-1mm} \includegraphics[width=0.32\textwidth]{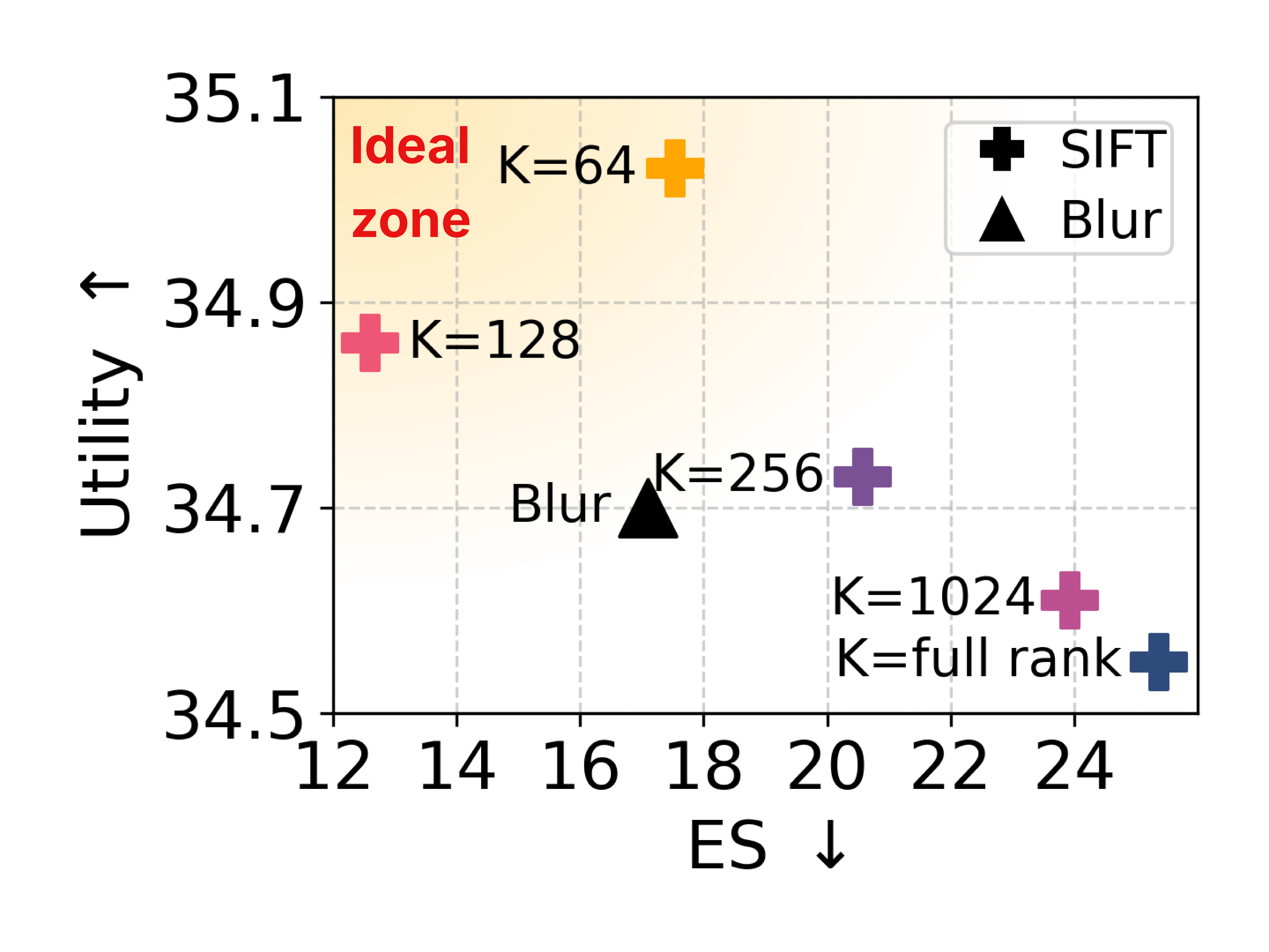}
     % \\
     %  {\scriptsize{(a) Intrinsic low-rank structure}} &  {\scriptsize{(b) Performance under different $K$}}
    \end{tabular} 
    \vspace*{-3mm}
\caption{{{Sensitivity analysis of SIFT with respect to the top-$K$ subspace dimension. Left: effective rank of momentum matrices. Right: unlearning and utility performance under varying $K$.}}}
    \label{fig:topk_unlearn}
   % \vspace*{-5mm}
%\end{wrapfigure}
\end{figure}

\noindent \textbf{A sensitivity analysis on unlearning vs. subspace dimension $K$.}
We next analyze the role of the top-$K$ spectral subspace in SIFT via (i) the intrinsic low-rank structure of momentum and (ii) the performance trade-off under varying $K$. 
Using SVD, we measure the cumulative energy 
$
\mathcal{E}(K)={\sum_{i=1}^{K}\sigma_i^2}/{\sum_{i=1}^{N}\sigma_i^2} $
and define the effective rank as the smallest $K$ such that $\mathcal{E}(K)\ge \alpha$.   \textbf{Fig.~\ref{fig:topk_unlearn}(Left)} shows the average effective rank of the momentum matrices associated with the updated MLP down-projection modules in the selected layers,
averaged across all training steps under SIFT.
As we can see, the effective rank is approximately
$K \approx 120$ at $\alpha = 99\%$ energy, far below the full dimension ($>4096$ for a 7B model), confirming a low-rank structure. 
Furthermore, \textbf{Fig.~\ref{fig:topk_unlearn}(Right)} shows a clear trade-off with respect to $K$: small $K$ (\textit{e.g.}, $64$) preserves utility but yields weak unlearning, while increasing $K$ improves unlearning, peaking at $K=128$. Larger $K$ (\textit{e.g.}, $> 256$) leads to degraded unlearning and utility due to over-expansion of the intervention subspace.

%\begin{wrapfigure}{r}{0.5\textwidth}

\subsection{Experiment results on safety alignment}
%\vspace{-5pt}
\label{sec:safety_exp}
% We perform safety alignment on Llama-2-7b to improve its safety behavior on the PKU-SafeRLHF dataset while preserving its general instruction-following performance on Alpaca.
In this application, we align the base model Llama-2-7b with safety requirements via preference optimization, which introduces an inherent trade-off between improving safety and preserving general utility. \textit{E.g.}, optimizing for safety often encourages refusal behaviors, potentially degrading performance on general instruction-following tasks. The specifications of the primary and constraint objectives, as well as the data--model--evaluation setups, are summarized in Table~\ref{tab:fg_spec} and Table~\ref{tab:setup}, respectively.

In \textbf{Table\,\ref{tab:safety_alignment}}, we report safety performance and general utility on LLaMA-2-7B, comparing SIFT with the original model and baseline methods.
% All metrics are higher-is-better.
% Consistent with the results of LLM unlearning in Table\,\ref{tab:sift_unlearn_main}, 
\textit{SIFT consistently outperforms all baselines on safety metrics while also improving utility.} Specifically, it achieves $42.8\%$ (SR), $31.0\%$ (JBB), and $56.0\%$ (WJ), surpassing BLUR ($35.8\%$, $29.0\%$, $54.4\%$) and others. Importantly, SIFT also improves utility over BLUR across all benchmarks, including MMLU (39.2\% vs.\ 37.4\%), GSM8K (11.6\% vs.\ 9.8\%), IFEval (32.7\% vs.\ 31.3\%), MNLI (40.5\% vs.\ 37.6\%), and MRPC (68.4\% vs.\ 65.9\%).
This advantage stems from their different mechanisms for handling objective conflict. BLUR removes gradient aligned with the safety objective, but also discards useful utility information. In contrast, SIFT performs localized spectral interference mitigation, preserving task-relevant components while removing only their interference. 
% \vspace{-8pt}
\begin{table}[htb]
\centering
\caption{{
Safety alignment performance across optimization methods. Higher scores on {Strong Reject}, {JBB}, and {Wild Jailbreak} indicate stronger safety, while higher scores on downstream tasks ({MMLU}, {GSM8K}, {IFEval}, {MNLI}, {MRPC}) indicate better utility. Evaluation metrics are specified in Table\,\ref{tab:setup} and results format follows Table\,\ref{tab:sift_unlearn_main}.
}}
\setlength{\tabcolsep}{4pt}
\resizebox{0.95\linewidth}{!}{
\begin{tabular}{l ccc|ccccc|c}
\toprule
\multirow{2}{*}{\textbf{Methods}} 
& \multicolumn{3}{c}{\textbf{Safety Performance} (\%)} 
& \multicolumn{5}{c}{\textbf{Utility Performance} (\%)}
& \multirow{2}{*}{\makecell{\textbf{Runtime} \\ (min)}} \\
\cmidrule(lr){2-4} \cmidrule(lr){5-9}
& SR $\uparrow$
& JBB $\uparrow$
& WJ $\uparrow$
& MMLU $\uparrow$
& GSM8K $\uparrow$
& IFEval $\uparrow$
& MNLI $\uparrow$
& MRPC $\uparrow$
& \\
\midrule
Base model & 19.5\scriptsize{$\pm 0.8$} & 14.0\scriptsize{$\pm 1.2$} & 47.6\scriptsize{$\pm 1.2$} & 41.3\scriptsize{$\pm 0.3$} & 14.7\scriptsize{$\pm 0.8$} & 34.2\scriptsize{$\pm 0.4$} & 42.8\scriptsize{$\pm 0.3$} & 69.5\scriptsize{$\pm 0.2$} & N/A \\ 
\midrule
AdamW & 33.2\scriptsize{$\pm 0.6$} & 28.0\scriptsize{$\pm 1.1$} & 52.8\scriptsize{$\pm 0.4$} & 36.6\scriptsize{$\pm 0.5$} & 8.4\scriptsize{$\pm 0.9$} & 31.9\scriptsize{$\pm 0.3$} & 36.2\scriptsize{$\pm 0.2$} & 62.7\scriptsize{$\pm 0.3$} & 110.3 \\
Muon & 36.4\scriptsize{$\pm 0.3$} & 26.0\scriptsize{$\pm 1.3$} & 52.0\scriptsize{$\pm 0.9$} & 36.1\scriptsize{$\pm 0.5$} & 9.5\scriptsize{$\pm 0.7$} & 30.4\scriptsize{$\pm 0.4$} & 32.8\scriptsize{$\pm 0.1$} & 63.3\scriptsize{$\pm 0.4$} & 191.2 \\
POME & 36.7\scriptsize{$\pm 0.6$} & 26.0\scriptsize{$\pm 1.1$} & 51.6\scriptsize{$\pm 1.2$} & 38.7\scriptsize{$\pm 0.3$} & 9.2\scriptsize{$\pm 0.5$} & 31.6\scriptsize{$\pm 0.5$} & 32.5\scriptsize{$\pm 0.2$} & 62.1\scriptsize{$\pm 0.5$} & 115.4 \\
BLUR & 35.8\scriptsize{$\pm 0.3$} & 29.0\scriptsize{$\pm 0.5$} & 54.4\scriptsize{$\pm 0.9$} & 37.4\scriptsize{$\pm 0.6$} & 9.8\scriptsize{$\pm 0.4$} & 31.3\scriptsize{$\pm 0.8$} & 37.6\scriptsize{$\pm 0.4$} & 65.9\scriptsize{$\pm 0.2$} & 114.7 \\
\rowcolor{LightCyan!50}
SIFT & 42.8\scriptsize{$\pm 0.6$} & 31.0\scriptsize{$\pm 0.4$} & 56.0\scriptsize{$\pm 1.0$} & 39.2\scriptsize{$\pm 0.6$} & 11.6\scriptsize{$\pm 0.5$} & 32.7\scriptsize{$\pm 0.5$} & 40.5\scriptsize{$\pm 0.4$} & 68.4\scriptsize{$\pm 0.6$} & 215.6 \\
\bottomrule
\end{tabular}
}
\label{tab:safety_alignment}
% \vspace{-10pt}
\end{table}

% This enables stronger safety alignment without sacrificing utility.

%\begin{figure}[htb]
% \begin{wrapfigure}{r}{0.45\textwidth}
%     \centering
%     \vspace*{-6mm}
%     \includegraphics[width=.93\linewidth]{figure/safety_heatmap_v2.pdf}
%     \vspace*{-5mm}
%     \caption{\footnotesize{Localization patterns of SIFT for safety alignment, similar to Fig.\,\ref{fig:motivation_gradient_alignment}.}}
%     \label{fig:safety_heatmap}
% %\end{figure}
% \vspace*{-5mm}
% \end{wrapfigure}

% In addition, SIFT for safety alignment exhibits a sparse localization pattern across both optimization steps and model components; see \textbf{Fig.~\ref{fig:safety_heatmap}} in \textbf{Appendix\,\ref{app:safety_k}}. Compared to the pattern observed in Fig.~\ref{fig:motivation_gradient_alignment} for LLM unlearning, safety alignment shows a distinctly different behavior. In particular, primary–constraint conflicts arise in more localized regions, \textit{e.g.}, middle layers and early optimization stages.
For safety alignment, SIFT shows sparse, localized intervention across steps and components (\textbf{Fig.~\ref{fig:safety_heatmap}}), distinct from LLM unlearning patterns (Fig.~\ref{fig:motivation_gradient_alignment}), with conflicts mainly in middle layers and early steps.
Furthermore, consistent with the low-rank observation in Fig.~\ref{fig:topk_unlearn}, we find that choosing $K=192$, aligned with the intrinsic effective rank, yields improved performance in both safety and utility; see \textbf{Fig.\,\ref{fig:topk_safety}}.
\subsection{Experiment results on text-to-speech adaptation}
%\vspace{-5pt}
\label{sec:speech_exp}
% We fine-tune GLM-4-Voice model on the \textit{interleaved} speech–text data constructed from ESNLI, COSE, and OpenBookQA (see \textbf{Appendix\,\ref{app:speech_data}} for details). Our goal is to jointly improve its generation quality on these tasks across both text and audio modalities while avoiding cross-modal forgetting \citep{cuervo2025closing, chen2024voicebench}.
In this application, we adapt the base model GLM-4-Voice to acquire dual text--speech generation capabilities on specific tasks, where the key challenge of model steering is to jointly improve generation across text and audio modalities without incurring cross-modal forgetting \citep{cuervo2025closing, chen2024voicebench}. We formulate this as a constrained optimization problem, as specified in Table~\ref{tab:fg_spec}. Specifically, we fine-tune the base model on \textit{interleaved} speech--text data derived from the training sets of ESNLI, COSE, and OpenBookQA.
The construction process of the interleaved speech--text data and an example can be found in \textbf{Appendix\,\ref{app:speech_data}}.
We evaluate both text and audio generation on their respective test sets; see Table~\ref{tab:setup} for the data--model--evaluation setups.

\begin{table}[htb]
% \vspace*{-5mm}
% \caption{\footnotesize{Test accuracy ($\uparrow$) of training speech LLM with interleaved speech--text data from ESNLI, COSE, and OpenBookQA. Test accuracy is reported under four input--output settings (\textit{e.g.}, ``Audio $\rightarrow$ Text'' denotes audio input and text output). Evaluation metrics follow Table~\ref{tab:setup}, and the result format follows Table~\ref{tab:sift_unlearn_main}.}}
\caption{Test accuracy ($\uparrow$) of the speech LLM trained on interleaved speech–text data constructed from ESNLI, COSE, and OpenBookQA under four input–output settings (e.g., “Audio → Text” denotes audio input and text output). Evaluation metrics and result format follow Tables 2 and 3, respectively.}
\label{tab:speechLLM}
\centering
% \small
% \footnotesize
\resizebox{1\linewidth}{!}{
\begin{tabular}{lccc|ccc|ccc|ccc|c}
\toprule
\multirow{2}{*}{\textbf{Methods}} 
& \multicolumn{3}{c}{\textbf{Audio → Text} (\(\uparrow\))} 
& \multicolumn{3}{c}{\textbf{Audio → Audio} (\(\uparrow\))} 
& \multicolumn{3}{c}{\textbf{Text → Text} (\(\uparrow\))} 
& \multicolumn{3}{c|}{\textbf{Text → Audio} (\(\uparrow\))} 
& \multirow{2}{*}{\makecell{\textbf{Runtime} \\ (min)}} \\
\cmidrule(lr){2-4} \cmidrule(lr){5-7} \cmidrule(lr){8-10} \cmidrule(lr){11-13}
& ESNLI & COSE & OpenBook
& ESNLI & COSE & OpenBook
& ESNLI & COSE & OpenBook
& ESNLI & COSE & OpenBook & \\
\midrule
Base 
& 27.2 \scriptsize{$\pm 1.1$} & 42.1 \scriptsize{$\pm 1.1$}& 12.6\scriptsize{$\pm 1.3$}
& 20.1 \scriptsize{$\pm 2.1$}  & 41.6 \scriptsize{$\pm 2.2$} & 11.2 \scriptsize{$\pm 2.4$}
& 60.6 \scriptsize{$\pm 0.5$}& 66.2 \scriptsize{$\pm 0.6$}& 54.7 \scriptsize{$\pm 0.8$}
& 42.8 \scriptsize{$\pm 1.5$}& 34.5  \scriptsize{$\pm 1.5$}& 43.4  \scriptsize{$\pm 1.8$} & N/A \\
\midrule
AdamW
& 70.3 \scriptsize{$\pm 1.0$} & 46.7 \scriptsize{$\pm 1.3$}  & 48.6 \scriptsize{$\pm 1.2$}
& 68.6 \scriptsize{$\pm 1.7$} & 48.1 \scriptsize{$\pm 2.2$}& 27.5 \scriptsize{$\pm 2.2$}
& 78.9 \scriptsize{$\pm 0.5$}& 74.1 \scriptsize{$\pm 0.6$} & 68.2 \scriptsize{$\pm 0.8$}
& 74.1 \scriptsize{$\pm 1.8$} & 52.9  \scriptsize{$\pm 1.9$} & 54.2  \scriptsize{$\pm 1.5$} & 44.6 \\
Muon
& 73.6 \scriptsize{$\pm 1.2$}  &  48.5  \scriptsize{$\pm 1.1$} & 53.4 \scriptsize{$\pm 1.2$}
& 71.5 \scriptsize{$\pm 2.4$} & 49.1 \scriptsize{$\pm 2.1$}& 27.3 \scriptsize{$\pm 2.5$}
& 79.9 \scriptsize{$\pm 0.5$} & 72.8 \scriptsize{$\pm 0.6$} & 64.4 \scriptsize{$\pm 0.9$}
& 74.3 \scriptsize{$\pm 1.6$}& 52.4 \scriptsize{$\pm 1.5$} & 53.8 \scriptsize{$\pm 1.9$} & 47.9 \\
POME
& 73.3 \scriptsize{$\pm 1.1$} & 47.2 \scriptsize{$\pm 1.4$} & 53.4 \scriptsize{$\pm 1.3$}
& 70.6 \scriptsize{$\pm 1.9$} & 48.2 \scriptsize{$\pm 2.1$}& 25.7 \scriptsize{$\pm 2.5$}
& 80.6 \scriptsize{$\pm 0.5$} & 72.3 \scriptsize{$\pm 0.5$}& 64.5 \scriptsize{$\pm 0.9$}
& 74.1 \scriptsize{$\pm 1.5$} & 52.2  \scriptsize{$\pm 1.6$}& 53.7  \scriptsize{$\pm 2.0$} & 24.7 \\
BLUR
& 69.6 \scriptsize{$\pm 1.2$} & 43.2  \scriptsize{$\pm 1.2$} & 47.1 \scriptsize{$\pm 1.3$}
& 68.6 \scriptsize{$\pm 2.4$} & 44.1 \scriptsize{$\pm 2.5$}& 20.4 \scriptsize{$\pm 2.3$}
& 80.1 \scriptsize{$\pm 0.5$} & 72.2 \scriptsize{$\pm 0.6$}& 63.1 \scriptsize{$\pm 0.8$}
& 77.4 \scriptsize{$\pm 1.9$}& 53.6  \scriptsize{$\pm 1.7$}& 51.5  \scriptsize{$\pm 1.6$} & 46.1\\
\rowcolor{LightCyan!50}
\textbf{SIFT}
& 77.4 \scriptsize{$\pm 1.1$} & 56.3 \scriptsize{$\pm 1.3$} & 57.1 \scriptsize{$\pm 1.2$}
& 77.1 \scriptsize{$\pm 2.4$} & 53.4 \scriptsize{$\pm 2.1$}& 29.5 \scriptsize{$\pm 2.3$}
& 80.4 \scriptsize{$\pm 0.5$} & 75.2 \scriptsize{$\pm 0.6$} & 64.1 \scriptsize{$\pm 0.9$}
& 79.6 \scriptsize{$\pm 1.6$}& 53.5  \scriptsize{$\pm 1.7$}& 54.3  \scriptsize{$\pm 1.9$} &  51.3\\
\bottomrule
\end{tabular}
}
\end{table}

% \noindent \textbf{Performance and insights.}
% In \textbf{Table~\ref{tab:speechLLM}}, we report test accuracy on ESNLI, COSE, and OpenBookQA across four cross-modal input--output settings, comparing SIFT with the base model GLM-4-Voice   and baselines (AdamW, Muon, POME, and BLUR). 
%

In \textbf{Table~\ref{tab:speechLLM}}, we report test accuracy across four input--output settings, comparing SIFT with the base model GLM-4-Voice 
and baseline methods.
As shown, \textit{SIFT consistently achieves superior performance across nearly all datasets and settings}. In particular, under \textit{audio-input settings}, SIFT significantly outperforms the second-best method (Muon), with average gains of $5.1\%$ on ``Audio $\rightarrow$ Text'' and $4.0\%$ on ``Audio $\rightarrow$ Audio'' across datasets.
These settings are more challenging due to the need for accurate semantic extraction from audio, where SIFT’s ability to mitigate cross-modal conflicts is especially beneficial.
In contrast, control 
intervention-present methods such as POME perform similarly to Muon, while 
BLUR consistently underperforms across most settings, suggesting that its projection discards gradient components critical for training speech LLMs.

\begin{figure}[htb]
%\begin{wrapfigure}{r}{0.4\textwidth}
    \centering
    \includegraphics[width=.45\linewidth]{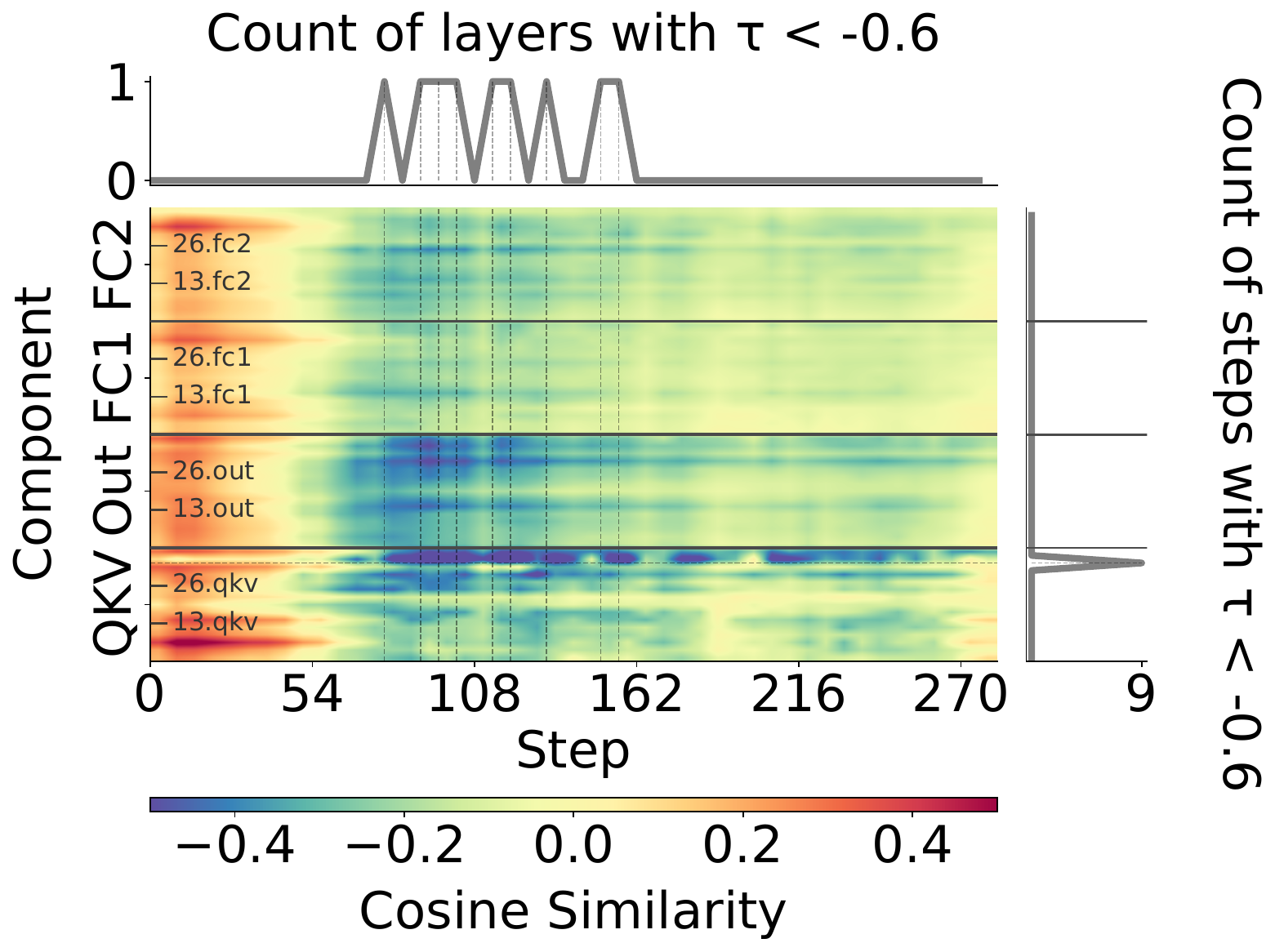}
    \caption{
     Localization patterns under the dataset ESNLI. 
    Components are grouped by Transformer layer: each layer comprises self-attention module (Q, K, V, Out) and feed-forward module (FC1, FC2). The presenting format follows Fig.\,\ref{fig:motivation_gradient_alignment}.
    }
    \label{fig:speech_qkv}
%\vspace*{-8mm}
%\end{wrapfigure}
\end{figure}

Another key observation is that, \textit{interference between the primary and constraint objectives is largely confined to the query, key, and value (QKV) matrices of self-attention layers}.
% Accordingly, SIFT adopts a localization scheme that selectively intervenes on these components for subspace control.
% \textbf{Fig.~\ref{fig:speech_qkv}} visualizes gradient misalignment (cosine similarity) across optimization steps (temporal) and model layers (spatial) on ESNLI. 
Similar 
to Fig.\,\ref{fig:motivation_gradient_alignment},
\textbf{Fig.~\ref{fig:speech_qkv}} visualizes gradient misalignment on ESNLI. 
% For ease of visualization, we group the same components across layers; in GLM-4-Voice, the Q, K, and V matrices are concatenated and shown as ``QKV''. 
As shown, negative cosine similarity is concentrated in the ``QKV'' regions, particularly in one layer (layer 38).
% indicating that conflicts are highly localized in these components.
It is also worth noting that, unlike Fig.~\ref{fig:topk_unlearn} and Fig.~\ref{fig:topk_safety}, SIFT for speech LLM adaptation (as well as the 
hallucination mitigation) does not benefit from reduced dimension $K$ in subspace control. Instead, using the full dimension yields better performance.

\subsection{Hallucination mitigation}
%\vspace*{-2mm}
\label{sec:hallucination_exp}

% We fine-tune  Llama-2-7B-Chat on RAGTruth \citep{niu2024ragtruth} to mitigate its word-level hallucinations. Model’s response may contain both hallucinated and non-hallucinated content; see \textbf{Appendix\,\ref{app:rag_exa}} for details. 
In this application, we fine-tune the base model Llama-2-7B-Chat to mitigate its word-level hallucinations. As illustrated in \textbf{Table~\ref{tab:examples_hallucination}},
% of \textbf{Appendix\,\ref{app:rag_exa}}
 the base model’s response may contain both hallucinated (highlighted in red) and non-hallucinated content. Our goal is to suppress hallucinated words via the unlearning objective while preserving non-hallucinated content through the standard training objective.  The specifications of the primary and constraint objectives, as well as the data--model--evaluation setups, are summarized in Table~\ref{tab:fg_spec} and Table~\ref{tab:setup}, respectively. 
We use RAGTruth \citep{niu2024ragtruth} for both training and evaluation (on its test sets).

\begin{table}[htb]
\centering
\caption{Performance of hallucination mitigation on RAGTruth. Hallucination rate measures the proportion of responses containing hallucinated content (judged by GPT-5.2); lower is better. 
}
\label{tab:hallucination}
% \small
\resizebox{.9\linewidth}{!}{
\begin{tabular}{lccccc c c}
\toprule
\multirow{2}{*}{\textbf{Methods}} 
& \multirow{2}{*}{\makecell{\textbf{Hallucination} \\ \textbf{Rate} (\(\downarrow\))}} 
& \multicolumn{5}{c}{\textbf{Utility Performance} (\(\uparrow\))} 
& \multirow{2}{*}{\makecell{\textbf{Runtime} \\ (min)}} \\
\cmidrule(lr){3-7}
& & MMLU & GSM8K & TruthfulQA & QNLI & MNLI & \\
\midrule
Base  model
& 73.2 \scriptsize{$\pm 1.6$}
& 46.5\scriptsize{$\pm 0.4$} & 20.4 \scriptsize{$\pm 1.1$} & 30.2 \scriptsize{$\pm 1.6$} & 68.7 \scriptsize{$\pm 0.6$} & 56.2 \scriptsize{$\pm 0.5$} & N/A \\
\midrule
AdamW
& 39.1 \scriptsize{$\pm 1.9$}
& 46.1 \scriptsize{$\pm 0.4$}& 17.4 \scriptsize{$\pm 1.1$} & 28.9 \scriptsize{$\pm 1.6$} & 68.3 \scriptsize{$\pm 0.6$} & 56.2\scriptsize{$\pm 0.5$} &  8.1\\
Muon
& 38.5 \scriptsize{$\pm 1.7$}
& 46.2 \scriptsize{$\pm 0.4$}& 17.8 \scriptsize{$\pm 1.1$} & 29.0 \scriptsize{$\pm 1.5$} & 68.2 \scriptsize{$\pm 0.6$} & 56.2 \scriptsize{$\pm 0.5$} & 12.0\\
POME
& 41.1 \scriptsize{$\pm 1.5$}
& 46.0 \scriptsize{$\pm 0.4$}& 17.8 \scriptsize{$\pm 1.1$} & 29.2 \scriptsize{$\pm 1.6$}& 68.2 \scriptsize{$\pm 0.6$}& 56.1 \scriptsize{$\pm 0.5$} & 10.6\\
BLUR
& 44.3 \scriptsize{$\pm 2.2$}
& 46.0 \scriptsize{$\pm 0.4$}& 15.4 \scriptsize{$\pm 1.0$} & 27.1 \scriptsize{$\pm 1.6$}& 68.0 \scriptsize{$\pm 0.6$}& 56.2 \scriptsize{$\pm 0.5$} & 8.6\\
\rowcolor{LightCyan!50}
\textbf{SIFT}
& 32.7 \scriptsize{$\pm 1.6$}
& 46.2 \scriptsize{$\pm 0.4$}& 17.8 \scriptsize{$\pm 1.1$} & 29.0 \scriptsize{$\pm 1.6$}& 68.2 \scriptsize{$\pm 0.6$} & 56.2 \scriptsize{$\pm 0.5$} &  26.4\\
\bottomrule
\end{tabular}
}
\end{table}

% \noindent \textbf{Performance and insights.}
% In \textbf{Table~\ref{tab:hallucination}}, we report hallucination reduction (Hallucination Rate) and model utility (\textbf{MMLU}, \textbf{GSM8K}, \textbf{TruthfulQA}, \textbf{QNLI}, \textbf{MNLI}) across optimization methods. 
In \textbf{Table~\ref{tab:hallucination}}, we report hallucination reduction and model utility across different optimizers. 
As shown, \textit{SIFT achieves the best trade-off between hallucination reduction and utility retention}.
To note, unlike other methods, BLUR exhibits notable drops on utility, \textit{e.g.}, GSM8K (15.4\%) and TruthfulQA (27.1\%), compared to SIFT (17.8\% and 29.0\%).  This is expected, as BLUR discards gradient components important for preserving utility.
% In terms of hallucination reduction, SIFT achieves the lowest rate (32.7\%), outperforming Muon (38.5\%) by a clear margin. And these gains do not compromise utility: SIFT achieves the best performance on MMLU, MNLI, and GSM8K, and ranks second on TruthfulQA and QNLI.
In terms of hallucination reduction, SIFT achieves the lowest rate (32.7\%), outperforming the next-best Muon (38.5\%) by a clear margin, without compromising utility relative to other baselines.
% Finally, {Table~\ref{tab:examples_hallucination}} shows that responses after SIFT-based updating remain coherent and meaningful, without producing 
% repetitive or degenerate outputs.
We also provide examples in  {Table~\ref{tab:examples_hallucination}} to show that the mitigated model using SIFT produces coherent and meaningful responses, without repetition or degeneration.

Similar to text-to-speech adaptation, interference in hallucination mitigation is largely confined to the QKV matrices of self-attention layers. Similar to \textbf{Fig.~\ref{fig:speech_qkv}}, \textbf{Fig.~\ref{fig:hallucination_qkv}} in \textbf{Appendix\,\ref{app:heat_map_hallu}} shows that the primary-constraint conflict is concentrated in the QKV regions, particularly in Layer 27.Q, Layer 27.K, and Layer 28.K between steps 170 and 280.

% \input{section/method}
% \input{section/method_SL}

% \clearpage
% \newpage

% \input{section/prelimary}

% \input{section/exp}

\section{Conclusion}
\label{sec:conclusion}

% In this work, we introduce SIFT, a subspace-control framework designed to resolve the fundamental optimization conflicts in constrained model steering. SIFT is motivated by the connection between one-shot model merging and the spectral optimization framework Muon, where gradient orthogonalization serves to eliminate interference between objectives and constraints.By analyzing how interference emerges both temporally and spatially within model updates, SIFT employs a localization scheme to selectively intervene in the optimization process.
% Extensive experiments across machine unlearning, safety alignment, text‑to‑speech adaptation, and hallucination mitigation demonstrate that SIFT reliably improves the trade‑off between preserving utility and satisfying constraint objectives.

% In this paper, we present SIFT, a subspace-control framework designed to resolve the fundamental optimization conflicts in constrained model steering. To develop this, we establish a novel, unexplored connection between the one-shot model merging and the gradient-orthogonalized optimizer Muon. Moreover, SIFT employs a localization scheme to selectively intervene in optimization based on the systematic, localized conflict between the primary objective and constraints.  
% We conduct extensive experiments on four impactful model steering applications: machine unlearning, safety alignment, text‑to‑speech adaptation, and hallucination mitigation.
% And we show that SIFT consistently achieves superior performance compared to both control-based and control-free baselines.
% \SL{[future work, limitations]}

We present SIFT, a subspace-control framework that resolves optimization conflicts in constrained model steering. We uncover a novel connection between one-shot model merging and the gradient-orthogonalized optimizer Muon, and design a localization scheme to intervene on systematic, localized conflicts between objectives and constraints. We evaluate SIFT on four model steering applications, including machine unlearning, safety alignment, text-to-speech adaptation, and hallucination mitigation, and show that it consistently outperforms both control-based and control-free baselines.
Since SIFT uses SVD for subspace construction, improving efficiency is an important direction for future work. Another interesting direction is to investigate whether subspace control can be extended to the constrained pre-training paradigm, beyond the post-training setting considered in this work.

% \clearpage
% \newpage

% \input{section/Limitation}

% \bibliography{refs/RA,refs/MU,refs/MU_SLiu}

\bibliography{main}
\bibliographystyle{iclr2025_conference}

\clearpage\newpage
\onecolumn
\section*{{Appendix}}
%\appendix
\setcounter{section}{0}
\setcounter{figure}{0}
\setcounter{table}{0}
\makeatletter 
\renewcommand{\thesection}{\Alph{section}}
\renewcommand{\theHsection}{\Alph{section}}
\renewcommand{\thefigure}{A\arabic{figure}} % Figure counter representation
\renewcommand{\theHfigure}{A\arabic{figure}} % Hyperref figure hyperlink hook
\renewcommand{\thetable}{A\arabic{table}}
\renewcommand{\theHtable}{A\arabic{table}}
\makeatother
\renewcommand{\thetable}{A\arabic{table}}
\setcounter{mylemma}{0}
\renewcommand{\themylemma}{A\arabic{mylemma}}
\setcounter{algorithm}{0}
\renewcommand{\thealgorithm}{A\arabic{algorithm}}
\setcounter{equation}{0}
\renewcommand{\theequation}{A\arabic{equation}}

\section{SIFT Algorithm}
\label{app:alg}

\textbf{Algorithm\,~\ref{algo:sift}} presents the detailed procedure of our proposed method SIFT. 
At a high level, our approach treats model parameters in a structured manner and performs optimization under a constrained subspace to mitigate interference between objectives.

% Specifically, we consider two objectives: a primary objective $f$ and a constraint objective $g$, whose gradients may exhibit misalignment during training. 
% To address this, our method dynamically identifies and operates within a subspace that preserves useful descent directions while suppressing conflicting components.

The algorithm proceeds in three main stages. 
First, we evaluate the alignment between the gradients of $f$ and $g$ for each parameter block, which serves as a signal for detecting potential interference. 
Second, when significant misalignment is detected, we activate a subspace control mechanism following \eqref{eq:expansion}--\eqref{eq:SIFT_update} that constructs a structured update direction by selectively filtering gradient components. 
Otherwise, standard Muon optimization \eqref{eq:Muon} is applied without modification.  Overall, this procedure enables \textit{localized and adaptive control} over the optimization process, allowing the model to balance objective alignment while avoiding unnecessary loss of useful gradient information.

\begin{algorithm}[htb]
   \caption{SIFT with primary objective $f$ and constraint objective $g$}
   \label{algo:sift}
   \small
   \begin{algorithmic}[1]
      \STATE \textbf{Input:} Subspace dimension $K$, misalignment threshold $\epsilon < 0$, total steps $T$, stepsizes $\{\eta_t\}_{t=0}^{T-1}$, and initialization $\btheta_0$
      \FOR{$t=0,\ldots,T-1$}
         \FOR{each parameter block $l=1,\ldots,L$}
            \STATE Compute alignment score $\tau_t^{(l)}=\frac{\nabla f(\btheta_t^{(l)})^\top \nabla g(\btheta_t^{(l)})}
            {\|\nabla f(\btheta_t^{(l)})\|_2\,\|\nabla g(\btheta_t^{(l)})\|_2}$, where  $\btheta_t^{(l)}$ denotes the $l$th block of $\btheta_t$
            % \[
            % \tau_t^{(l)}=\frac{\nabla f(\btheta_t^{(l)})^\top \nabla g(\btheta_t^{(l)})}
            % {\|\nabla f(\btheta_t^{(l)})\|_2\,\|\nabla g(\btheta_t^{(l)})\|_2}.
            % \]
            \IF{$\tau_t^{(l)} < \epsilon$}
               \STATE Update $\btheta_{t+1}^{(l)}$ using SIFT via \eqref{eq:expansion}--\eqref{eq:SIFT_update} 
               {\hfill // with subspace control}
            \ELSE
               \STATE Update $\btheta_{t+1}^{(l)}$ using standard Muon via \eqref{eq:Muon}
            \ENDIF
         \ENDFOR
        % \STATE Aggregate $\{\btheta_{t+1}^{(l)}\}_{l=1}^L$ into $\btheta_{t+1}$
      \ENDFOR
      \STATE \textbf{Return:} $\btheta_T$
   \end{algorithmic}
\end{algorithm}

\begin{figure}[htp]
    \centering
    \includegraphics[width=0.5\linewidth]{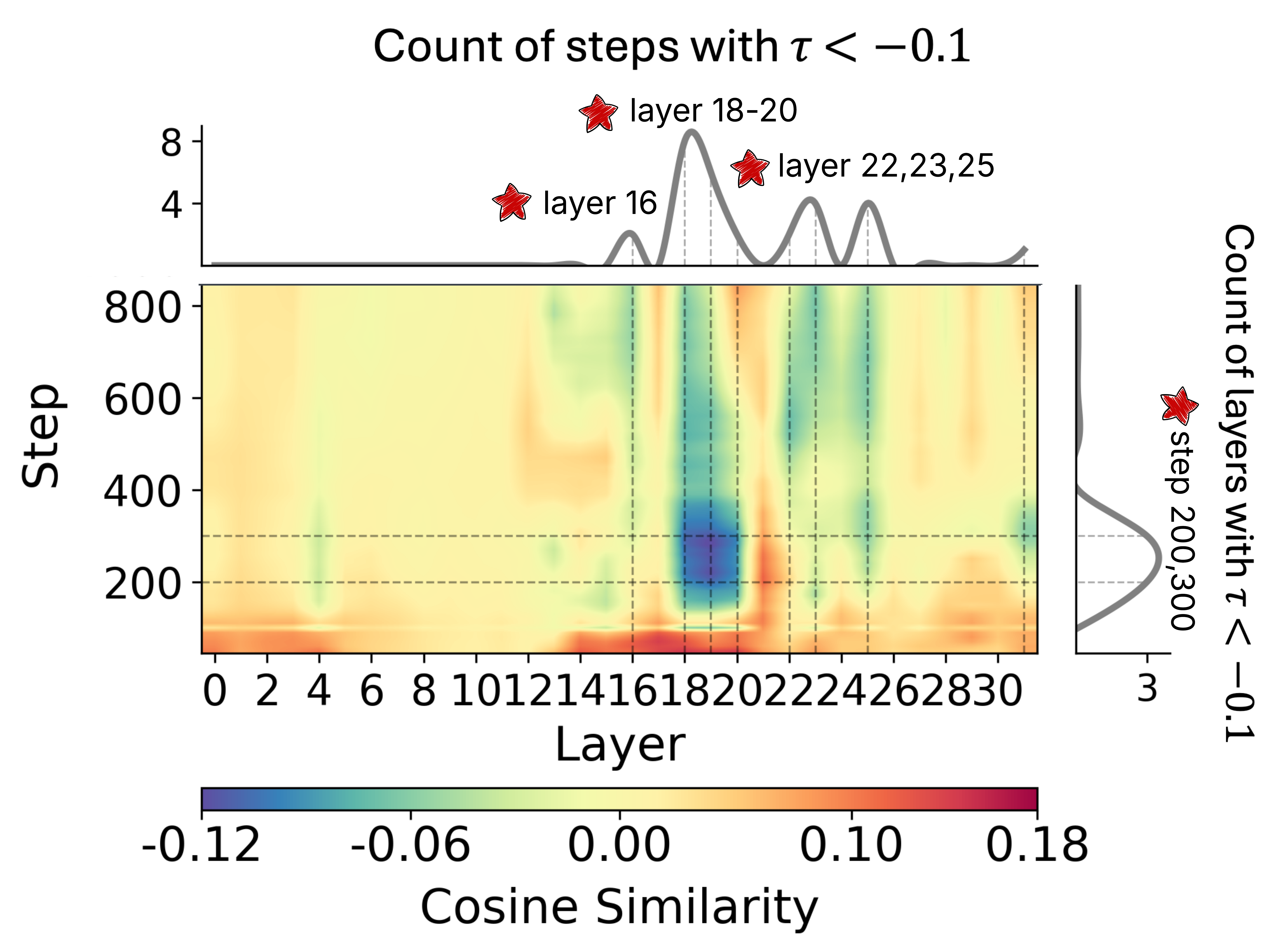}
    % \vspace*{-5mm}
    \caption{{Localization patterns of SIFT for safety alignment, similar to Fig.\,\ref{fig:motivation_gradient_alignment}.}}
    \label{fig:safety_heatmap}
\end{figure}

\section{Localization Patterns of SIFT for Safety Alignment}
\textbf{Fig.\,\ref{fig:safety_heatmap}} presents the gradient alignment pattern under the safety alignment setting, following the same visualization protocol as Fig.~\ref{fig:motivation_gradient_alignment}. 
In contrast to the unlearning case, where RMU operates on a subset of layers, SIFT in safety alignment is applied across all layers.
As we can see, SIFT exhibits sparse and localized intervention patterns across both optimization steps (\textit{temporal}) and model components (\textit{spatial}). 
Notably, the conflict regions are primarily concentrated in middle layers and occur at early training stages, which differs from the unlearning setting where conflicts tend to appear in higher layers.

\section{Sensitivity Analysis on Safety Alignment vs. Subspace Dimension $K$}
\label{app:safety_k}

\begin{figure}[htb]
    \centering
    \vspace*{-1mm}
    \begin{tabular}{cc}
     %\hspace*{-5mm} 
     \includegraphics[width=0.39\textwidth]{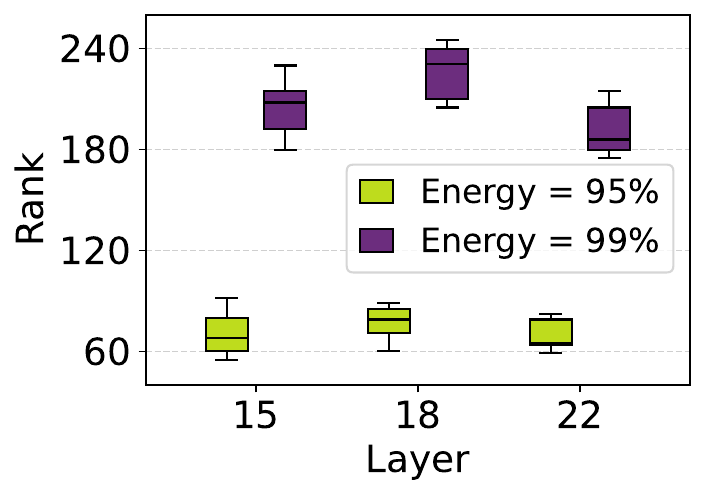}    &  
     %\hspace*{-5mm} 
     \includegraphics[width=0.39\textwidth]{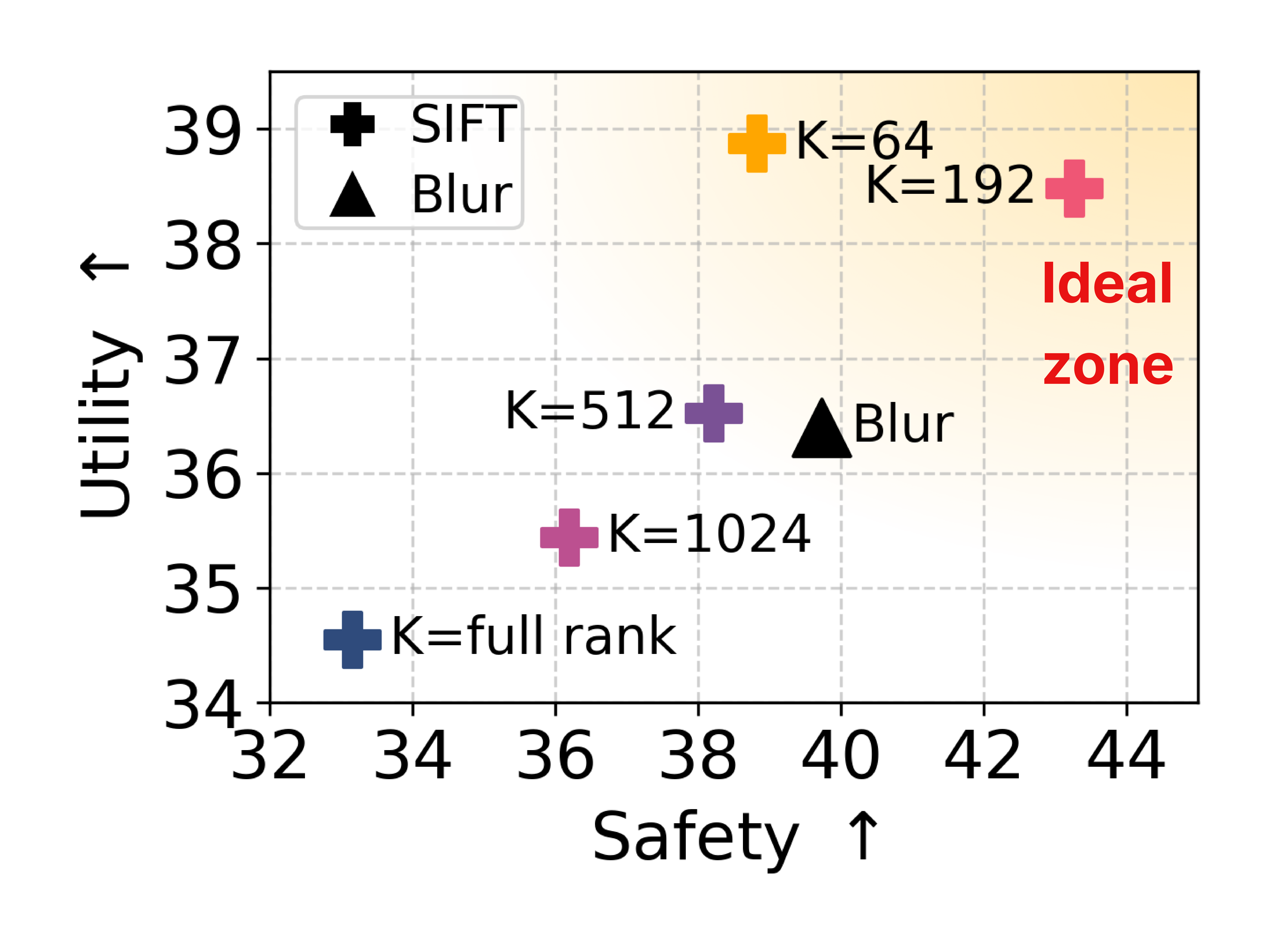}\\
    \end{tabular} 
    \vspace*{-3mm}
    \caption{{\footnotesize{Sensitivity analysis on subspace dimension $K$ in SIFT for safety alignment. (\textit{Left}) The effective rank of the momentum matrices computed across SIFT-localized layers (15, 18, and 22) under different energy thresholds.
 (\textit{Right}) Performance of SIFT under varying $K$, where Safety denotes the average over all safety metrics, and Utility denotes the average over all utility metrics in Table\,\ref{tab:safety_alignment}.}   }}
    \label{fig:topk_safety}
    \vspace*{-3mm}
%\end{wrapfigure}
\end{figure}

Similar to Fig.~\ref{fig:topk_unlearn}, we analyze the role of the top-$K$ spectral subspace in \textbf{Fig.~\ref{fig:topk_safety}} of SIFT under the safety alignment setting from both structural and performance perspectives.
Specifically, we examine (i) the intrinsic low-rank structure of momentum and (ii) the trade-off between safety and utility under varying $K$. \textbf{Fig.~\ref{fig:topk_safety}(Left)} shows the effective rank of the momentum matrices computed in layers 15, 18, and 22, averaged across training steps.
Consistent with the unlearning setting, the momentum exhibits a clear low-rank structure, where a relatively small $K \approx 200 $ captures the majority of spectral energy. Furthermore, \textbf{Fig.~\ref{fig:topk_safety}(Right)} illustrates the sensitivity of SIFT to different choices of $K$.
We observe a similar trade-off pattern: smaller $K$ (e.g., $64$) tends to preserve utility but yields limited safety improvement, while moderate $K$ (e.g., $192$) achieves the best balance.
In contrast, larger $K$ (e.g., $512$, $1024$, or full-rank) leads to performance degradation in both safety and utility, suggesting that overly expanding the intervention subspace introduces unnecessary interference. Overall, these results further validate that an appropriately chosen low-dimensional spectral subspace is critical for achieving effective and stable safety alignment.

\section{Interleaved Speech–Text Data Construction}
\label{app:speech_data}

Following \citep{zeng2024scaling, nguyen2025spirit}, we construct interleaved speech–text data separately from the text training corpora of COSE, e-SNLI, and OpenBookQA.
% and fine-tune models on each dataset individually. 
The construction procedure follows \citep{zeng2024scaling}. Specifically, we first convert text responses into speech using the Coqui TTS API \citep{coqui_tts_2021_v14}, and then tokenize the resulting audio with the GLM-4-Voice speech tokenizer to obtain speech token sequences. We construct interleaved sequences by alternating 13 text tokens with 26 speech tokens, following the standard output format of GLM-4-Voice.
\textbf{Fig.\,\ref{fig:speech_text_interleaved}} provides  an example of such interleaved data constructed from COSE's training set.

\begin{figure}[htbp]
\centering
\begin{tcolorbox}[
  colback=codebg,
  colframe=codeborder,
  boxrule=0.4pt,
  arc=1pt,
  left=6pt,right=6pt,top=6pt,bottom=6pt,
  width=0.9\textwidth   
]
{\small
\ttfamily
\texttokens{13} Answer: D. outside. Explanation: billy is an animal, 
\boaa\ \tokens{26} <|audio\_15358|><|audio\_9903|>...<|audio\_4037|><|audio\_7780|> \eoaa
\texttokens{13} but he is allergic to trees. He hates them very much. Still, he wants to have a picnic and can't 
\boaa\ \tokens{26} <|audio\_4914|><|audio\_5137|>...<|audio\_329|><|audio\_2429|> \eoaa
\texttokens{left} stand outside. \boaa\ \tokens{left} <|audio\_508|>\\<|audio\_9443|>...<|audio\_1089|><|audio\_2360|>\eoaa
}
\end{tcolorbox}
\vspace*{-8pt}
\caption{\footnotesize{Example of interleaved speech–text data constructed from COSE's training set.}}
\label{fig:speech_text_interleaved}
\end{figure}

\section{Example Responses from the RAGTruth Dataset}
\label{app:rag_exa}
We present an example of word-level LLM hallucination for an input query sampled from the RAGTruth summarization task in Table~\ref{tab:examples_hallucination}. The original model’s response may contain both hallucinated content (highlighted in red) and non-hallucinated content. Our objective is to suppress hallucinated words through the unlearning objective while preserving the non-hallucinated content.
Furthermore, the response from the mitigated model using SIFT remain coherent and meaningful, without exhibiting repetitive or degenerate outputs.

\begin{table}[htb]
\centering
\vspace{-5pt}
\caption{\footnotesize{
Example responses from the base model and the SIFT-updated model for an input query sampled from the RAGTruth dataset. Red text indicates hallucinated content.
}}
\vspace{5pt}
\label{tab:examples_hallucination}
\scalebox{0.9}{
\begin{tabular}{c|c}
\toprule[1pt]
\midrule
{\textbf{Input query}} & 
\parbox[c]{10cm}{
Summarize the following news within 161 words: …Five people were infected and three died in the past year in Kansas from listeria that might be linked to Blue Bell Creameries products, according to the CDC...} \\
\midrule
{\textbf{Original Model } }  & 
\parbox[c]{10cm}{
... This is the third time Blue Bell has taken action due to listeria contamination, and the company is cooperating with investigations. No illnesses have been reported directly linked to the contaminated ice cream, \textcolor{red}{but five people in Kansas have died from listeriosis in the past year after consuming Blue Bell products.}} \\
\midrule
{\textbf{Mitigated model}} 
& 
\parbox[c]{10cm}{
...This recall follows past listeria outbreaks in Kansas and Texas, where five people were infected and three died, some after consuming milkshakes made with Blue Bell ice cream. Blue Bell is cooperating with authorities, emphasizing safety, and other Blue Bell products are not affected.}
\\
\midrule
\bottomrule[1pt]
\end{tabular}
}
\end{table}

\section{Localization Patterns of SIFT for Hallucination Mitigation}
\label{app:heat_map_hallu}
\begin{figure}[htp]
    \centering
    \includegraphics[width=0.5\linewidth]{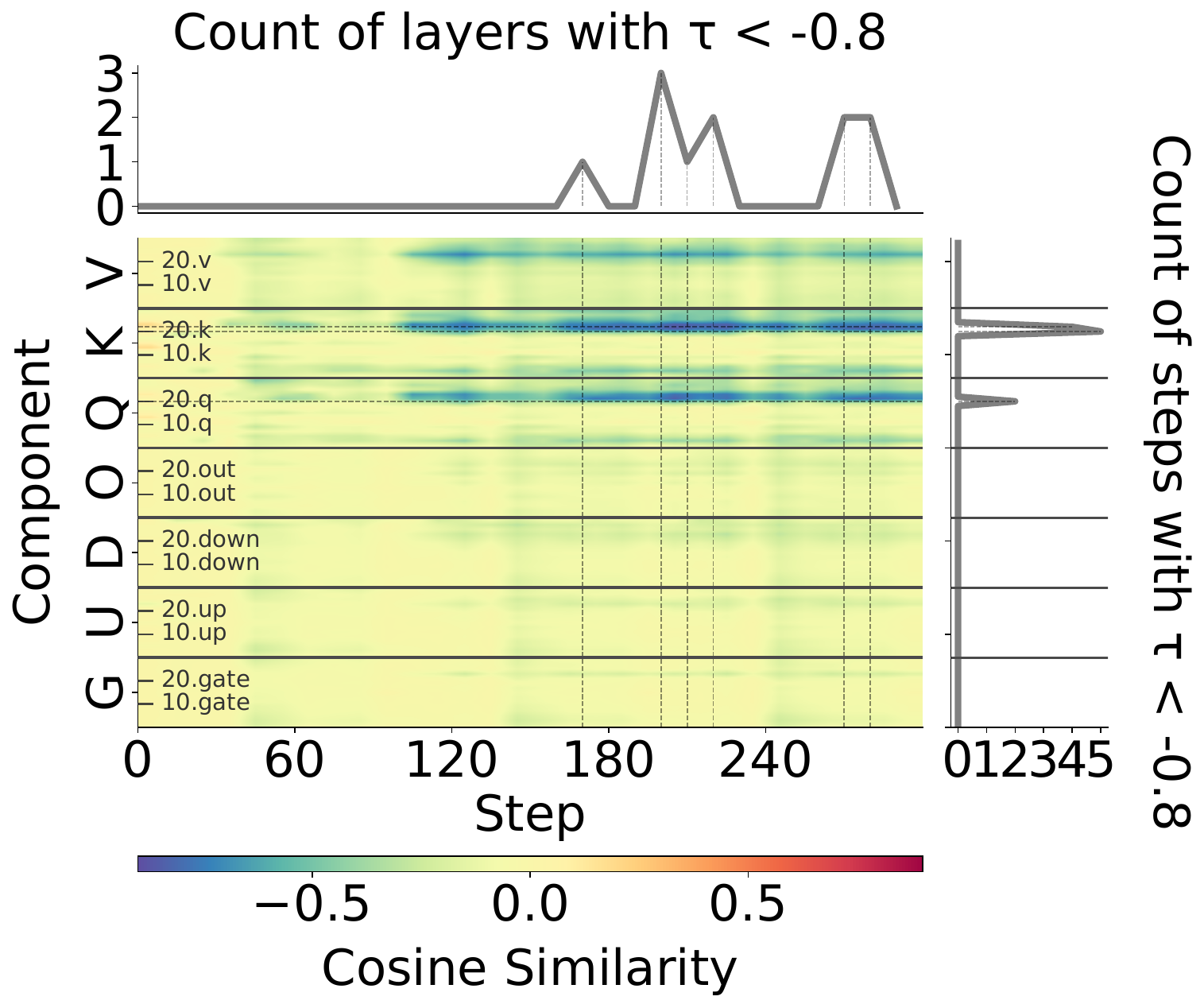}
    % \vspace*{-5mm}
    % \caption{\footnotesize{
    %  Localization patterns of SIFT for hallucination mitigation (similar format to Fig.~\ref{fig:motivation_gradient_alignment}). ``Q'', ``K'', and ``V'' denote the query, key, and value projections; ``O'' denotes the attention output projection; and ``G'', ``U'', and ``D'' denote the feed-forward network’s gating, up-, and down-projections.
    % }}
 \caption{In each Transformer layer, “Q”, “K”, and “V” denote the query, key, and value projection matrices, and “O” denotes the output projection of the self‑attention module. “G”, “U”, and “D” denote the gating, up‑projection, and down‑projection components of the feed‑forward network.}   
    \label{fig:hallucination_qkv}
\end{figure}

% Similar to \textbf{Fig.~\ref{fig:motivation_gradient_alignment}}, 
% \textbf{Fig.~\ref{fig:safety_heatmap}} and
% \textbf{Fig.~\ref{fig:speech_qkv}}, \textbf{Appendix\,\ref{app:heat_map_hallu}} 
Similar to \textbf{Fig.~\ref{fig:speech_qkv}}, \textbf{Fig.~\ref{fig:hallucination_qkv}} shows the gradient misalignment across optimization steps and model layers on RAGTruth. As observed, the negative cosine similarity is primarily concentrated in the QKV regions, particularly in Layer 27.Q, Layer 27.K, and Layer 28.K between steps 170 and 280. This pattern is consistent with the text-to-speech adaptation setting, where interference between the primary and constraint objectives during hallucination mitigation is largely localized to the QKV projections of the self-attention layers.

% \bibliography{refs/MU,refs/MU_SLiu,refs/PO}
% \bibliographystyle{iclr2025_conference}

\clearpage
\newpage

\end{document}